%% file: main_camera_ready.tex
\documentclass{article} 
\usepackage{iclr2025_conference,times}

\input{math_commands.tex}

\usepackage[colorlinks = true,
            linkcolor = black,
            urlcolor  = blue,
            citecolor = black,
            anchorcolor = blue]{hyperref}
\usepackage{url}

\usepackage{booktabs}       
\usepackage{amsfonts}       
\usepackage{nicefrac}       
\usepackage{microtype}      
\usepackage{xcolor}         

\usepackage{wrapfig}
\usepackage{subcaption}
\usepackage[normalem]{ulem}
\usepackage{tikz}

\iclrfinalcopy

\newcommand{\posteriors}{\texttt{posteriors}}
\newcommand{\mat}[1]{\mathrm{#1}}

\DeclareMathOperator{\gauss}{\mathbf{N}}

\DeclareMathOperator{\real}{\mathbf{R}}
\DeclareMathOperator{\bigOh}{\mathcal{O}}
\DeclareMathOperator{\F}{\mat{F}}
\DeclareMathOperator{\D}{\mat{D}}
\DeclareMathOperator{\Q}{\mat{Q}}
\DeclareMathOperator{\T}{\mathcal{T}}

\title{Scalable Bayesian Learning with \lowercase{\posteriors{}}}

\author{%
  Samuel Duffield\thanks{Correspondence to: \texttt{sam@normalcomputing.ai}}\\
  Normal Computing \\
  \And
  Kaelan Donatella\\
  Normal Computing\\
  \And
  Johnathan Chiu\\
  Normal Computing\\
  \AND
  Phoebe Klett\\
  Normal Computing\\
  \And
  Daniel Simpson\\
  Normal Computing
}

\begin{document}

\maketitle

\begin{abstract}
  Although theoretically compelling, Bayesian learning with modern machine learning models is computationally challenging since it requires approximating a high dimensional posterior distribution. In this work, we (i) introduce \posteriors{}, an easily extensible PyTorch library hosting general-purpose implementations making Bayesian learning accessible and scalable to large data and parameter regimes; (ii) present a tempered framing of stochastic gradient Markov chain Monte Carlo, as implemented in \posteriors{}, that transitions seamlessly into optimization and unveils a minor modification to deep ensembles to ensure they are asymptotically unbiased for the Bayesian posterior, and (iii) demonstrate and compare the utility of Bayesian approximations through experiments including an investigation into the cold posterior effect and applications with large language models.

    \begin{center}
        \posteriors{}: \href{https://github.com/normal-computing/posteriors}{\texttt{github.com/normal-computing/posteriors}}
    \end{center}
  
\end{abstract}

\section{Introduction}\label{sec:intro}

\begin{figure}
    \resizebox{\textwidth}{!}{
        \input{figs/bayes_benefits_schematic.tex}
    }
    \caption{\textbf{Pictorial representation of the benefits of Bayesian learning.} Left: Averaging over multiple plausible fits to the data improves out-of-distribution generalisation. Center: Adapting to new online data without forgetting previous data. Right: Decomposing predictive uncertainty, with epistemic uncertainty providing an improved indicator for out-of-distribution detection.}\label{fig:bayes_benefits}
\end{figure}
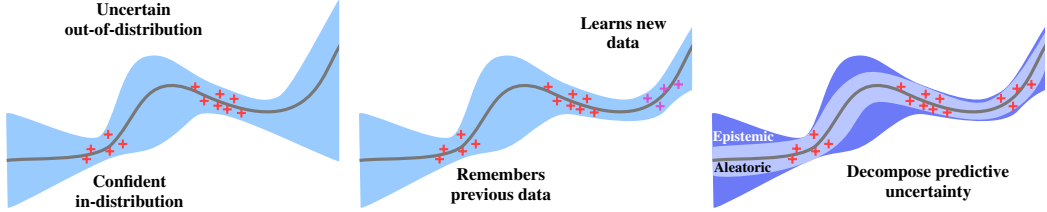

Bayesian learning is a framework for updating and inferring unknown model parameters in the presence of data. It is based on Bayes' theorem:
\begin{equation}\label{eq:bayes}
    p(\theta \mid y_{1:N}) \propto \underbrace{p(\theta) p(y_{1:N} \mid \theta)}_\text{offline} \propto \underbrace{p(\theta \mid y_{1:N-1}) p(y_N \mid \theta)}_\text{online},
\end{equation}
which provides a principled way to update our beliefs about the parameters $\theta \in \real^d$ as data is observed once $y_{1:N}$ or sequentially $y_N$. Bayesian learning represents uncertainty about the parameters using the posterior distribution $p(\theta \mid y_{1:N})$, which combines prior beliefs $p(\theta)$ with the likelihood or generating process $p(y \mid \theta)$ for the data. This contrasts with optimization approaches, which store a single point estimate of the parameters $\theta^* = \argmax_\theta p(y_{1:N} \mid \theta)$. In our view, for modern machine learning models and pipelines, Bayesian learning offers three key advantages, which are visualized in Figure~\ref{fig:bayes_benefits}: (a) improved generalization and out-of-distribution predictions, (b) coherent online learning of new information (without catastrophic forgetting), and (c) the ability to decompose predictive uncertainty into aleatoric (natural or explained uncertainty in the data) and epistemic (unexplained uncertainty that diminishes with more training data) components. It's important to note that in the more classical setting of statistical models with interpretable parameters, there are additional benefits to be gained from Bayesian learning - these include propagation of uncertainty \citep{duffield2024state} and model validation \citep{gelman2020bayesian}. In this work, we introduce \posteriors{}, a python package designed to be minibatch-first and easily applicable to large-scale models with PyTorch \citep{paszke2019pytorch} and Hugging Face \citep{Wolf_Transformers_State-of-the-Art_Natural_2020} whilst also being compatible with classical Bayesian models through composition with Pyro (\citealt{bingham2019pyro}, Appendix~\ref{sec:PI}).

Generalization is a key metric for machine learning models, quantifying the ability of the model to perform tasks beyond its training set. Bayesian predictions offer a compelling approach to mitigate overfitting by averaging over a posterior distribution of plausible models rather than relying on a single fit. In various settings, Bayesian methods have demonstrated improved out-of-distribution performance \citep{neal2012bayesian, blundell2015weight, liu2020simple}. 

Another vital consideration for modern machine learning pipelines is that of online or continual learning, where the model continuously updates and learns from new data in real-time without needing to be retrained from scratch. Bayes' theorem (\eqref{eq:bayes}) provides a coherent framework for online learning, where new data can be sequentially incorporated into the model by using the posterior distribution as the prior on receipt of the next data batch. When implemented exactly, this is equivalent to learning all data $y_{1:N}$ at once \eqref{eq:bayes}, and all data points are exchangeable and treated identically. This contrasts to naive online optimization, which typically results in the catastrophic forgetting \citep{goodfellow2013empirical, kirkpatrick2017overcoming} of previously learnt tasks.

In addition, it is essential for statistical models to be auditable and reliable, especially when they are met with data that is far outside the training distribution. With an optimization approach, the model is only able to provide predictions in the form of total uncertainty for the predictive distribution over test inputs $x^*$, via the entropy $H[p(y \mid x, \theta^*)] := - \E_{p(y \mid x, \theta^*)}[\log p(y \mid x, \theta^*)]$. This is in contrast to Bayesian methods, which can decompose the total uncertainty (TU) of the posterior predictive distribution $p(y \mid x) = \E_{p(\theta \mid y_{1:N})}[p(y \mid x, \theta)]$ into two components \citep{wimmer2023quantifying, sale2023second, hofman2024quantifying}:
\begin{align}
    \text{TU} := H[p(y \mid x)] , \quad
    \text{AU} := \E_{p(\theta \mid y_{1:N})} [H[p(y \mid x, \theta)]] ,  \quad
    \text{EU} &:= \text{TU} - \text{AU}. \label{eq:uncs}
\end{align}
Aleatoric uncertainty (AU) represents natural uncertainty in the data, such as synonyms or sentence starts in natural language generation. Epistemic uncertainty (EU) represents uncertainty in the model parameters and diminishes with more data (as $p(\theta \mid y_{1:N}) \to \delta(\theta \mid \theta^*)$ as $N \to \infty$) and therefore is a better indication of model uncertainty in predictions. This decomposition is crucial for understanding model behaviour and making informed decisions in high-stakes applications \citep{kendall2017uncertainties}, although is not unique to Bayesian models but rather the broader concept of second-order uncertainty \citep{osband2024epistemic, sale2023second, hofman2024quantifying}.

In this work, we concisely survey techniques for Bayesian learning with a view to scalability in both large data and large parameter regimes (with an extensive survey found in Appendix~\ref{sec:survey}). We then make the following contributions:
\begin{enumerate}
    \item We introduce \posteriors{}; a functional, minibatch-first PyTorch \citep{paszke2019pytorch} package for scalable and extensible Bayesian learning.
    \item We formalize gradient descent methods as the low-temperature limit of stochastic gradient Markov chain Monte Carlo (SGMCMC) methods, thus implying a simple modification of deep ensembles to parallel SGMCMC that ensures convergence to the Bayesian posterior.
    \item We provide experiments demonstrating the broad utility of \posteriors{} and the accessibility of the aforementioned benefits. We investigate the generalization performance of a host of approximate Bayesian methods and the prominence of the \textit{cold posterior effect} \citep{wenzel2020good, izmailov2021bayesian}, composing \posteriors{} with LoRA \citep{hu2021lora} to mitigate forgetting in a continual learning task with Llama 2 \citep{touvron2023llama} and out-of-distribution detection via epistemic uncertainty with Llama 3 \citep{llama3modelcard}. 
\end{enumerate}
We conclude with a discussion of related work and an outlook on future avenues for research and impact.

\section{Bayesian learning}\label{sec:bayes-learn}

We now review the most prominent Bayesian learning methods for large-scale problems, all of which are implemented in \posteriors{}.  An in-depth survey with full details and discussions on caveats and computational bottlenecks can be found in Appendix~\ref{sec:survey}.

A \textbf{Laplace approximation} is motivated by a second-order Taylor expansion around the mode of the posterior distribution $\theta_\text{MAP} = \argmax_\theta p(\theta \mid y_{1:N})$. This gives a Gaussian approximation $p(\theta \mid y_{1:N}) \approx \gauss (\theta \mid \theta_\text{MAP}, \hat{\Sigma})$. In the supervised learning setting, the covariance $\hat{\Sigma}$ is typically chosen \citep{daxberger2021laplace} to be an approximation to the Fisher information matrix $\F(\theta) = \E_{p(x, y\mid \theta)}[\nabla^2_\theta \log p(\theta \mid x, y)]$ which is intractable as we do not have access to the true distribution over inputs $p(x)$. Instead, two approximations \citep{kunstner2019limitations} are used based on the training data $\hat{p}(x, y) = N^{-1}  \sum_{i=1}^N \delta(x, y \mid x^i, y^i)$. The first we denote the \textit{empirical Fisher} $\F_\text{E}(\theta) = \E_{\hat{p}(x, y)} [\nabla_\theta\log p(\theta \mid y) \nabla_\theta\log p(\theta \mid y)^\top]$ and is easy to compute. The second we denote the \textit{conditional Fisher} $\F_\text{C}(\theta) = \E_{\hat{p}(x) p(y \mid x, \theta)} [\nabla_\theta\log p(\theta \mid y) \nabla_\theta\log p(\theta \mid y)^\top]$ and can be computed efficiently for common neural network likelihoods via an equivalence with the Generalized Gauss-Newton matrix (GGN, \citealt{martens2020new}) with further details in Appendix~\ref{subec:laplace}.

In its most common form \textbf{variational inference} (VI, \citealt{blei2017variational}) also forms a Gaussian approximation $p(\theta \mid y_{1:N}) \approx q_\phi(\theta) = \gauss (\theta \mid \mu, \Sigma)$. However, VI directly learns the mean and covariance $\phi = (\mu, \Sigma)$ through an optimization routine to minimize the Kullback-Leibler objective $\text{KL}[q_\phi(\theta) \mid \mid p(\theta \mid y_{1:N})]$, whose gradient can be approximated with Monte Carlo samples and tricks including reparameterzation \citep{rezende2014stochastic} and stick-the-landing \citep{roeder2017sticking}.

An important caveat to note for Gaussian approximations is that the full covariance matrix requires $\bigOh(d^2)$ memory and often $\bigOh(d^3)$ runtime for the required inversion and sampling operations. In the large parameter regime, this is prohibitive, and further approximations are required \citep{mackay1992practical, ritter2018scalable, daxberger2021laplace}. Additionally, we can reduce memory at inference time by constructing uncertainty directly in function space by \textbf{linearizing the forward model} \citep{foong2019between, immer2021improving}. If the likelihood has the form $p(y \mid x, \theta) = p(y \mid f_\theta(x))$ and we have a posterior approximation $\gauss(\theta \mid \mu, \Sigma)$ then the linearized forward distribution becomes $\gauss (f(x) \mid f_\mu(x) , \nabla_\theta f_\mu(x)^\top \Sigma \nabla_\theta f_\mu(x))$ where $\nabla_\theta f_\mu(x)$ is the Jacobian of $f_\theta(x)$ evaluated at $\mu$. This avoids having to sample many large parameter vectors.

\textbf{Stochastic gradient Markov chain Monte Carlo} (SGMCMC, \citealt{ma2015complete, nemeth2021stochastic}) instead forms a Monte Carlo approximation $p(\theta \mid y_{1:N}) \approx N^{-1}\sum_{k=1}^K \delta(\theta \mid \theta_k)$. The samples $\{\theta_k\}_{k=1}^K$ are typically collected by (approximately) evolving a stochastic differential equation (SDE) - see Section~\ref{sec:sg} and Appendix~\ref{sec:sgmcmc} - and collecting samples along the trajectory. Traditional full-batch Markov chain Monte Carlo (MCMC) schemes \citep{andrieu2003introduction} use a Metropolis-Hastings step to ensure convergence of the Monte Carlo approximation is preserved through discretization. In the minibatch setting, the Metropolis-Hastings ratio cannot be estimated easily. Instead, suppose we have access to an unbiased stochastic gradient. Then, as noted in \cite{welling2011bayesian}, the bias in the trajectory decreases quadratically $\bigOh(\epsilon_t^2)$ for learning rate $\epsilon_t$. Therefore, adopting a suitably decreasing learning rate schedule $\sum \epsilon_t^2 < \sum \epsilon_t = \infty$ \citep{robbins1951stochastic} will ensure an asymptotically unbiased Monte Carlo approximation.


\begin{figure}[!tbp]
  \centering
  \begin{minipage}[b]{0.35\textwidth}
    \centering
    \includegraphics[width=\textwidth]{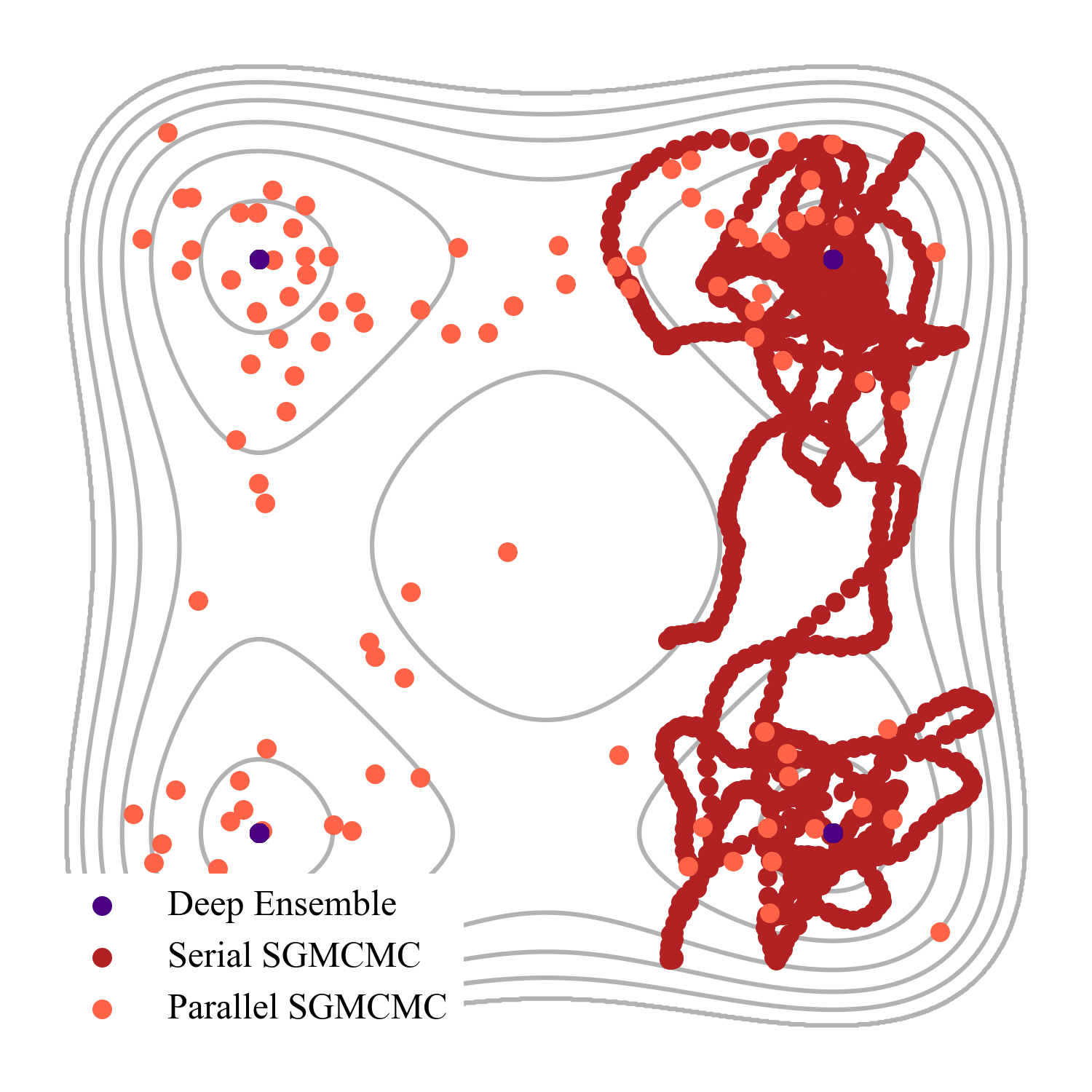}
    \caption{\textbf{Trajectories of various sampling methods} for a toy multimodal posterior. Deep ensemble concentrates on the modes, serial SGMCMC struggles to transfer between modes, parallel SGMCMC combines the benefits of both.}\label{fig:schematic}
  \end{minipage}
  \hfill
  \begin{minipage}[b]{0.60\textwidth}
    \centering
    \includegraphics[width=0.8\textwidth]{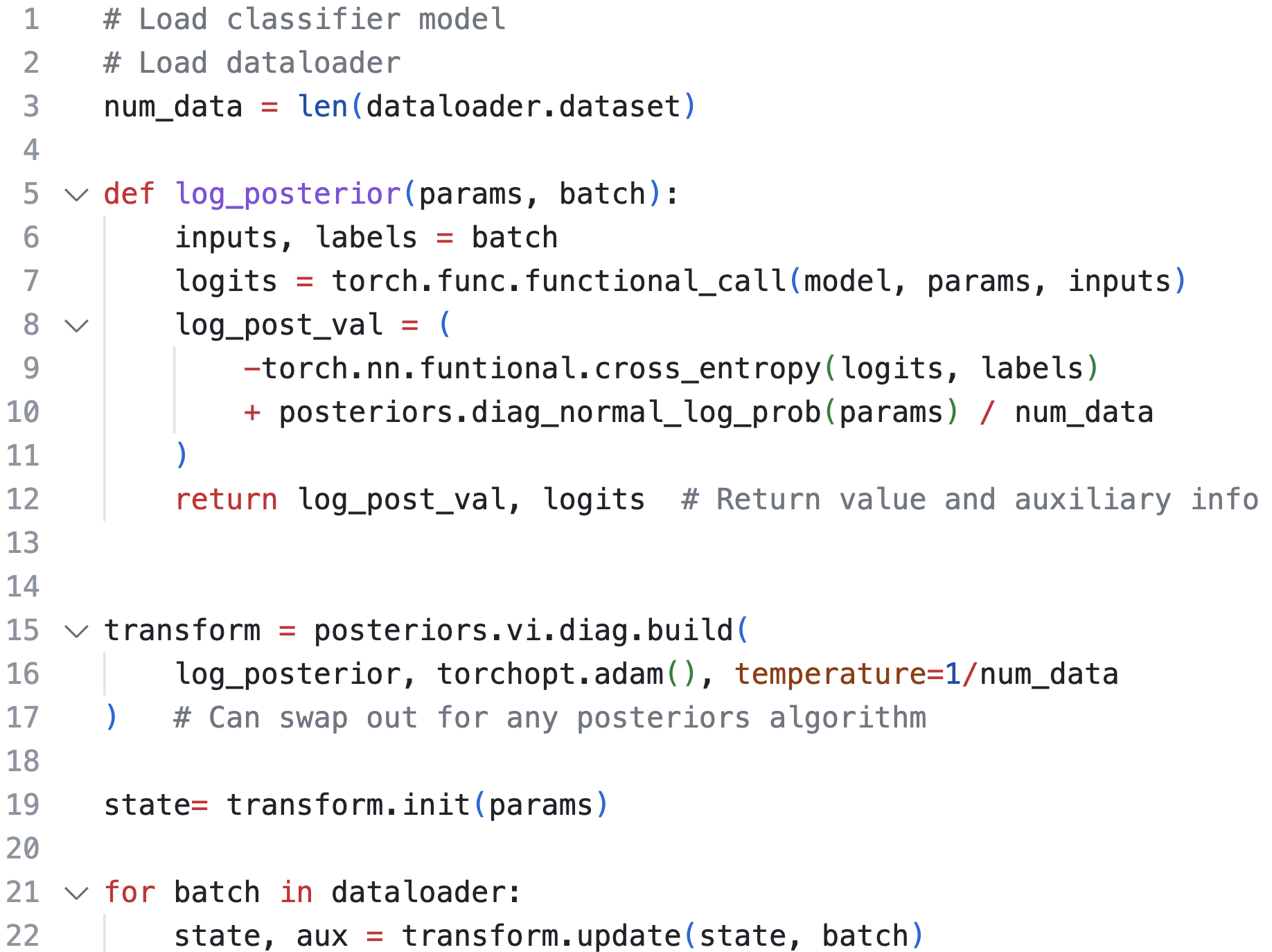}
    \caption{\posteriors{} \textbf{code snippet} to train a classifier with variational inference. \posteriors{} recommends normalising the log posterior across the batch with scale independent of batch size or $N$. Scaling the prior and temperature by $N^{-1}$ ensures the posterior is still correctly targeted.}\label{fig:code}
  \end{minipage}
\end{figure}

\section{\posteriors{}: Uncertainty quantification with PyTorch}\label{sec:posteriors}

\posteriors{} is designed to be a comprehensive library for uncertainty quantification in deep learning models. The key features outlining the \posteriors{} philosophy are:
\begin{itemize}
    \item \textbf{Composability}: \posteriors{} is written in PyTorch \citep{paszke2019pytorch} and has compatibility with a broad range of tools including the Llama \citep{touvron2023llama} and Mistral \citep{jiang2023mistral} models, \texttt{lightning} \citep{Falcon_PyTorch_Lightning_2019} for convenient logging and device management, optimizers from \texttt{torchopt} \citep{Ren_TorchOpt_2023} and probabilistic programs from \texttt{pyro} \citep{bingham2019pyro}, see Appendix~\ref{sec:PI}.
    
    \item \textbf{Extensible}: The transform framework (Figure~\ref{fig:code}) adopted by \posteriors{} is very general and allows for the easy adoption of new algorithms. Additionally, \posteriors{} supports arbitrary likelihoods rather than being restricted to hard-coded regression or classification as is common in other libraries \citep{daxberger2021laplace, detommaso2023fortuna}.
    
    \item \textbf{Functional}: \posteriors{} adopts a functional API via PyTorch's \citep{paszke2019pytorch} functional module. As championed by the JAX \citep{jax2018github} ecosystem, the functional approach makes for code that is easier to test and compose with other functions. Importantly for \posteriors{}, functional programming is also closer to the mathematical description, which is particularly useful for Bayesian modelling.
    
    \item \textbf{Scalable}: Unlike most common Bayesian software \citep{carpenter2017stan, cabezas2024blackjax, bingham2019pyro}, \posteriors{} is \textbf{minibatch-first}, thus allowing for efficient computation in large datasets. Additionally, flexible sub-spacing is provided for scaling to large models.
    
    \item \textbf{Swappable}: The transform framework (Figure~\ref{fig:code}) allows users to switch between approaches seamlessly, allowing them to experiment and find the best method for the use case.
\end{itemize}

This combination of features makes \posteriors{} unique amongst the PyTorch ecosystem which hosts the majority of the popular pretrained Hugging Face models \citep{Wolf_Transformers_State-of-the-Art_Natural_2020}.

\subsection{Supported Methods and Complexities}

All methods discussed in Section~\ref{sec:bayes-learn} are included in \posteriors{}. Flexible Laplace approximations are supported based on both the empirical Fisher and conditional Fisher (via the GGN matrix, see Section~\ref{sec:bayes-learn} and Appendix~\ref{sec:survey}). Laplace approximations rely on an optimization routine to find the MAP followed by a $O(bd)$ time and memory operation (for batch size $b$) to construct diagonal Fisher information matrices which ultimately require $O(d)$ memory. For very large models this can be a memory bottleneck meaning the batchsize may need to be reduced for the Laplace construction in comparison to the optimization (although the Laplace construction requires only a single epoch).

Variational inference for fitting a Gaussian distribution with diagonal covariance costs $O(kd)$ time and memory per iteration where $k$ is the number of samples from the variational distribution which is often set to 1 although larger $k$ reduces the variances of the gradient estimation. Dense covariance versions of both Laplace and VI are also supported but require $O(d^2)$ memory and $O(d^3)$ for sampling, however, this can still be useful for machine learning models when combined with subspace approaches \citep{peft, daxberger2021laplace}.

As we will see in Section~\ref{sec:sg}, SGMCMC is closely related to stochastic gradient descent and has the same per-iteration complexity. Although, a key difference is the final output is an ensemble of parameters rather than the single point estimate. As such memory requirements are $O(Md)$ where $M$ is the number of output samples, as with deep ensembles \citep{lakshminarayanan2017simple}. Further, as a suitable decay in autocorrelation is desired between samples the number of iterations is typically longer for SGMCMC than optimization, although this can be parallelized, see Section~\ref{subsec:psgmcmc}.

In addition, \posteriors{} hosts extended Kalman methods \citep{ollivier2018online} (which has the same complexity as VI, \citealt{zhang2018noisy}), unified optimization via torchopt \citep{Ren_TorchOpt_2023} and has a roadmap of methods to be added within the general framework including SVGD \citep{liu2016stein} and more advanced SGMCMC \citep{ma2015complete, zhang2020csgmcmc}. There is also a host of general-purpose utility functions not found elsewhere in the PyTorch ecosystem. This includes linearized forward calls, Fisher/GGN matrices, their diagonal counterparts, fast Fisher/GGN vector products and the conjugate gradient method, which provides the tools to implement Hessian-free optimization \citep{martens2010deep} and experiment with other second-order methods \citep{titsias2024optimal, duffield2022quasi}.

\subsection{Limitations}\label{subsec:limitations}
\posteriors{} supports arbitrary likelihoods and loss functions; as a result methods that utilise the specific form of the loss function (normally as classification or regression), e.g. the deterministic last layer approach in \cite{harrison2024variational} or the probit approximations from \cite{daxberger2021laplace}, are currently out of scope.

Similarly, the current methods in \posteriors{} do not assume any specific structure of the model or its parameters, see Figure~\ref{fig:code}. However, this makes it more challenging to support model-aware techniques such as DVI~\citep{wu2018deterministic} or K-FAC~\citep{martens2015optimizing}, although the benefits of such approaches may warrant an inclusion in the library going forward. Specifically for K-FAC, the Fisher information matrix is assumed to exhibit a local block structure that admits cheap inversion and sampling via Kronecker-factors. Initially, K-FAC was only developed for feedforward layers \citep{martens2015optimizing}, before being extended to convolutional~\citep{grosse2016kronecker} and recurrent~\citep{martens2018kronecker} networks. More recently, it was generalized to weight-sharing networks \citep{eschenhagen2024kronecker} (including transformers and graph neural networks). Developing a flexible K-FAC submodule within \posteriors{}' functional framework is a high priority. Such a module should be suitably model-agnostic and capable of being applied across various inference methods (e.g., optimization, Laplace, VI). This advancement would enable scalable inference with non-diagonal Fisher information matrices, broadening the library's capabilities.

Convergence diagnostics (e.g. \citealt{cowles1996markov}) represent a key component of MCMC workflow, however are not included directly in \posteriors{} due to the easy composition with Pyro's \citep{bingham2019pyro} extensive suite of convergence diagnostics. Similarly, \posteriors{} is an inference library rather than a probabilistic programming language, therefore does not include tools such as distribution definitions which can be utilized through composition with Pyro as demonstrated in Appendix~\ref{sec:PI}. Finally, we opt not to include a \texttt{fit} convenience function \citep{chollet2015keras, daxberger2021laplace} in favour of the more flexible and transparent \texttt{update} framework in Figure~\ref{fig:code}.

\section{Stochastic gradient descent, SGMCMC and deep ensembles}\label{sec:sg}

This section describes a tempered version of SGMCMC that introduces an additional temperature parameter that allows us to view SGMCMC as a generalization of stochastic gradient descent (SGD) and variants. This viewpoint has guided the implementation of \posteriors{} allowing users access to SGMCMC within a familiar functional framework of popular optimization libraries such as torchopt \citep{Ren_TorchOpt_2023} and optax \citep{deepmind2020jax}, thus making Bayesian learning scalable and maximally accessible to the broader machine learning community. In the second half of this section, we show how this points to a small modification of deep ensembles \citep{lakshminarayanan2017simple} to ensure convergence to the posterior, and therefore how \posteriors{} can be used to easily convert a deep ensemble pipeline into one of (parallel) Bayesian learning. Comprehensive details on SGMCMC are found in Appendix~\ref{subsec:SGMCMC}.

The tempered connection between SGMCMC and SGD has been discussed in previous works for specific optimizers/samplers e.g. \cite{wenzel2020good}, however here we give a complete characterization of the connection based on the framework in \cite{ma2015complete} which dictates the implementations in \posteriors{}. Additionally, to our knowledge, the corollary unifying deep ensembles and parallel SGMCMC is novel.

The minibatch-first nature of \posteriors{} alongside the reframing of stochastic gradient optimization as SGMCMC results in an implementation with access to both approaches within a unified syntax as well as enabling the codesign of new algorithms, all within a familiar optimization-style framework.


\subsection{A unified framework for SGD and SGMCMC}\label{subsec:SGMCMCopt}

Consider targeting a tempered target distribution $\pi_{\T}(z) \propto \pi(z)^\frac{1}{N\T}$ for temperature $\T \in [0, \infty)$, which can be considered as a synthetic hyperparameter controlling the variance of $\pi_{\T}(z)$ \citep{wenzel2020good, ding2014bayesian}. Here $z = (\theta, \omega)$ where $\omega$ are any auxiliary parameters such as momenta \citep{chen2014stochastic} or thermostat \citep{ding2014bayesian}. The distribution $\pi(z)$ is thus an extended target with $p(\theta \mid y_{1:N})$ marginal in $\theta$.

An SDE is invariant for $\pi_{\T}(z)$ if and only if it has the following form \citep{ma2015complete}
\begin{align}
    dz &= \T^{-1}[\D(z) + \Q(z)] N^{-1}\nabla \log \pi(z) dt + \nabla \cdot [\D(z) + \Q(z)] dt + \sqrt{2 \D(z)} dw, \nonumber \\
    \implies dz &= [\D(z) + \Q(z)] N^{-1}\nabla \log \pi(z) dt + \T\nabla \cdot [\D(z) + \Q(z)] dt + \sqrt{2 \T \D(z)} dw,
    \label{eq:tempsde}
\end{align}
where the second line comes from rescaling $\D(z), \Q(z) \to \T\D(z), \T\Q(z)$ \footnote{Equivalently we can rescale $dt \to \T dt$ as Brownian motion ($dw$) scales with $\sqrt{t}$, see (1.27) in \cite{pavliotis2014stochastic}}.

Setting $\mathcal{T}=N^{-1}$ regains the original target, but conveniently, the scale of the drift term $N^{-1}\nabla \log \pi(z)$ no longer grows with $N$ as $\log \pi(\theta) = \log p(\theta) + \sum_{n=1}^N \log p(y_n \mid \theta) + \text{const}$. This normalization of the loss or log posterior is the convention for optimization and results in more consistent learning rate selection across problems and is recommended in \posteriors{} - see Figure~\ref{fig:code}.

Intuitively, the low-temperature limit $\T \to 0$ converts the target distribution $\pi(z)$ into one with all probability mass concentrated on (local) maxima and transparently the SDE \eqref{eq:tempsde} into an ODE
\begin{align} \label{eq:sgd}
    \T \to 0 \implies dz = [\D(z) + \Q(z)] N^{-1} \nabla \log \pi(z) dt,
\end{align}
which represents exactly a variant of continuous time gradient ascent on $N^{-1} \log \pi(z)$.


All SGMCMC algorithms in \posteriors{} feature the temperature $\mathcal{T}$ as a parameter with $\mathcal{T}=0$ resulting in a valid SGD variant (specific examples can be found in Appendix~\ref{sec:sgmcmc-opt}). The extensibility and strict support for minibatches allow a unified \posteriors{}' implementation and new procedures immediately apply to both SGMCMC and SGD (e.g. step-size adaptation, which has in classical Bayesian statistics utilized Metropolis acceptance rates which do not apply to SGMCMC).

\subsection{Parallel SGMCMC (as Bayesian deep ensembles)}\label{subsec:psgmcmc}

Deep ensembles \citep{lakshminarayanan2017simple} represent a simple to implement and parallelizable route to uncertainty quantification via multiple runs of gradient descent with different random seeds (for initialization and batch shuffling). Thus, the arguments above point to the formulation of Bayesian deep ensembles, enabled by \posteriors{}, that converts the gradient descent ODE \eqref{eq:sgd} typically used for deep ensemble training into a Bayesian version that is stationary for the posterior distribution by setting $\T = N^{-1}$, i.e. adding back the noise term in \eqref{eq:tempsde}. Assuming $\D$ and $\Q$ are chosen to be independent of $z$ this becomes
\begin{align} \label{eq:bayes_de}
    dz = [\D + \Q] N^{-1} \nabla \log \pi(z) dt + \sqrt{2N^{-1}\D} dw.
\end{align}
The core difference between SGMCMC and the Bayesian deep ensemble or parallel SGMCMC implementation above is that in SGMCMC, samples are collected from a single trajectory, whereas in parallel SGMCMC, multiple trajectories are simulated with only the last sample being retained (although multiple samples per trajectory can also be saved \citep{margossian2021nested}). Parallel SGMCMC provides an asymptotically unbiased approximation to $\pi(z)$ and therefore $p(\theta~\mid~y_{1:N})$ as long as the criteria discussed in Appendix~\ref{subsec:SGMCMC} are met for a single chain.


We note that in the large data regime, the additional noise term in \eqref{eq:bayes_de} versus \eqref{eq:sgd} becomes very small, and so existing deep ensembles \citep{lakshminarayanan2017simple} can already be seen as approximating the posterior in some sense (with a small bias).

The functional framework of \posteriors{} (in combination with \texttt{torch.func}) allows easy parallelization of the above unified SGMCMC and SGD algorithms into a parallel SGMCMC implementation that generalizes deep ensembles with minimal code change.

Figure~\ref{fig:schematic} visualizes the behaviour of the techniques discussed in this section for a two-dimensional multi-modal double well posterior. We see clearly that although a deep ensemble finds all modes, it offers no within-mode uncertainty. In contrast, serial SGMCMC explores well within modes but struggles to transition between modes. Parallel SGMCMC combines the benefits of both methods to provide a more representative posterior approximation. In more realistic high dimensional, large $N$ settings runtime and memory requirements become a key consideration. Serial SGMCMC requires less total computation since it runs only a single trajectory. However, the trajectory must be significantly longer and cannot be parallelized; therefore, serial SGMCMC is perhaps not as well suited to modern computational architectures as the deep ensemble or parallel SGMCMC approaches. However, parallel SGMCMC naturally shares many of the caveats of deep ensembles. Notably, the resource requirements are significant to parallelize trajectories, although we note the opportunity to collect multiple samples per trajectory \citep{margossian2021nested}. Another point of note is that communication costs between cores are expensive and hyperparameter tuning may require alternative approaches to the serial setting and typically need more care (than optimization) to preserve the correct target distribution across trajectories.

\section{Experiments}\label{sec:experiments}

We now present experiments investigating the Bayesian learning benefits motivated in Section~\ref{sec:intro}. The functional and swappable nature of \posteriors{} makes these experiments particularly accessible across a range of Bayesian learning algorithms. Notably, its scalability and composability with the PyTorch ecosystem allow us to apply Bayesian learning to a host of pre-trained HuggingFace models \citep{Wolf_Transformers_State-of-the-Art_Natural_2020}. Reproducible code for all experiments (and more) can be found in the \href{https://github.com/normal-computing/posteriors/tree/main/examples}{examples folder} of the \posteriors{} repository.


\subsection{Cold Posterior Effect}\label{subsec:cpe}

This experiment fully utilizes \posteriors{} swappability between algorithms to investigate the generalization capabilities of Bayesian learning. As seen in Figure~\ref{fig:code}, within a \posteriors{} implementation the majority of the code is algorithm agnostic and we can easily swap approaches by only changing the build of the \texttt{transform} object and its configuration parameters. Extensive experimental details can be found in Appendix~\ref{sec:details_cpe}.


The cold posterior effect \citep{wilson2020bayesian, wenzel2020good, aitchison2020statistical, izmailov2021bayesian} is a phenomenon that has been observed for approximate Bayesian machine learning models where increased predictive performance has been observed when targeting $p_T(\theta \mid y_{1:N}) \propto p(\theta \mid y_{1:N})^{\frac1T}$ (as discussed in Section~\ref{sec:sg} with $\T = TN^{-1}$) for $T<1$, which appears at odds with the Bayesian paradigm \citep{aitchison2020statistical}.

\begin{figure}[ht]
    \centering
    \begin{subfigure}[b]{0.32\textwidth}
        \includegraphics[width=\textwidth]{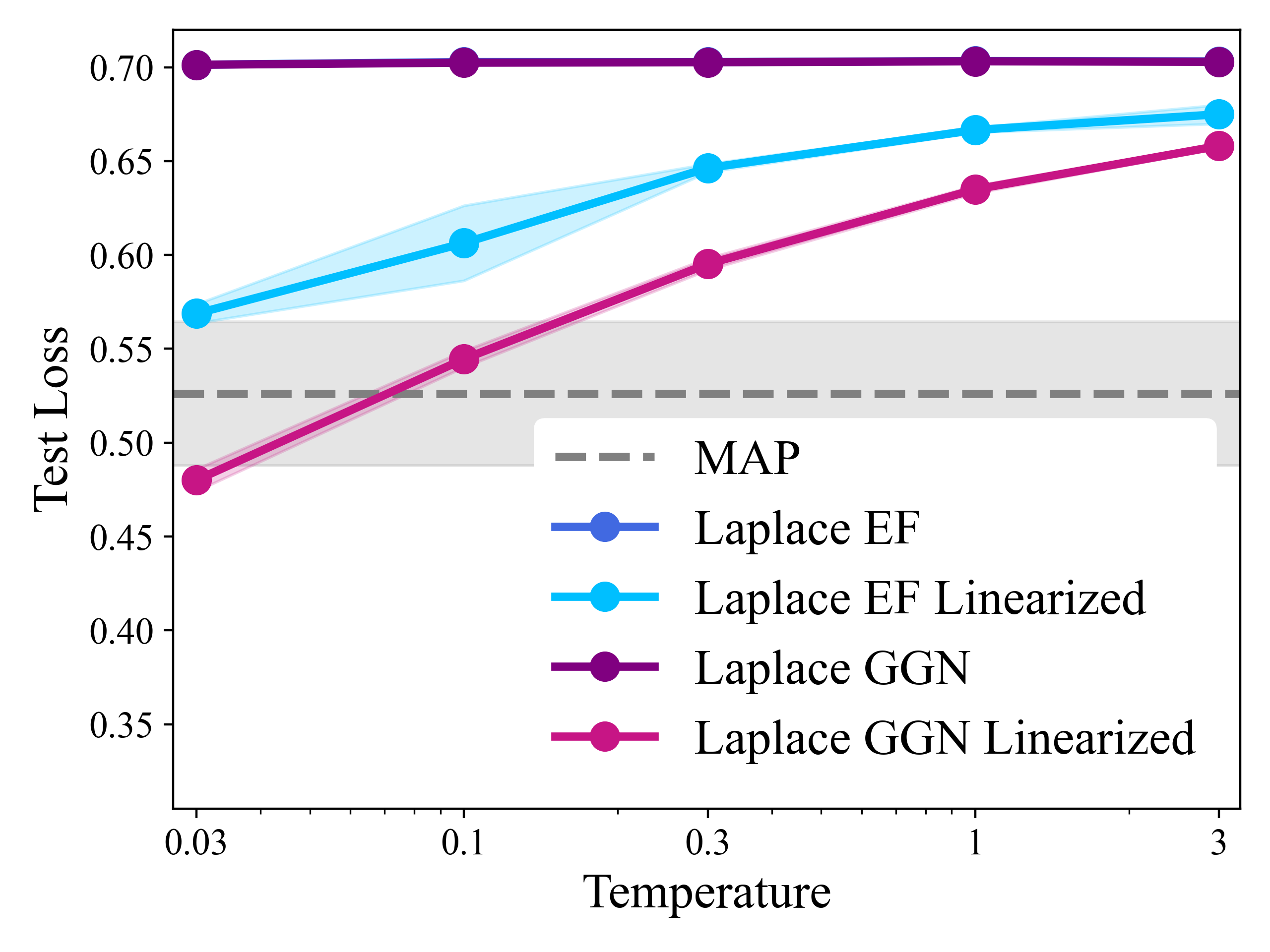}
    \end{subfigure}
    \hfill 
    \begin{subfigure}[b]{0.32\textwidth}
        \includegraphics[width=\textwidth]{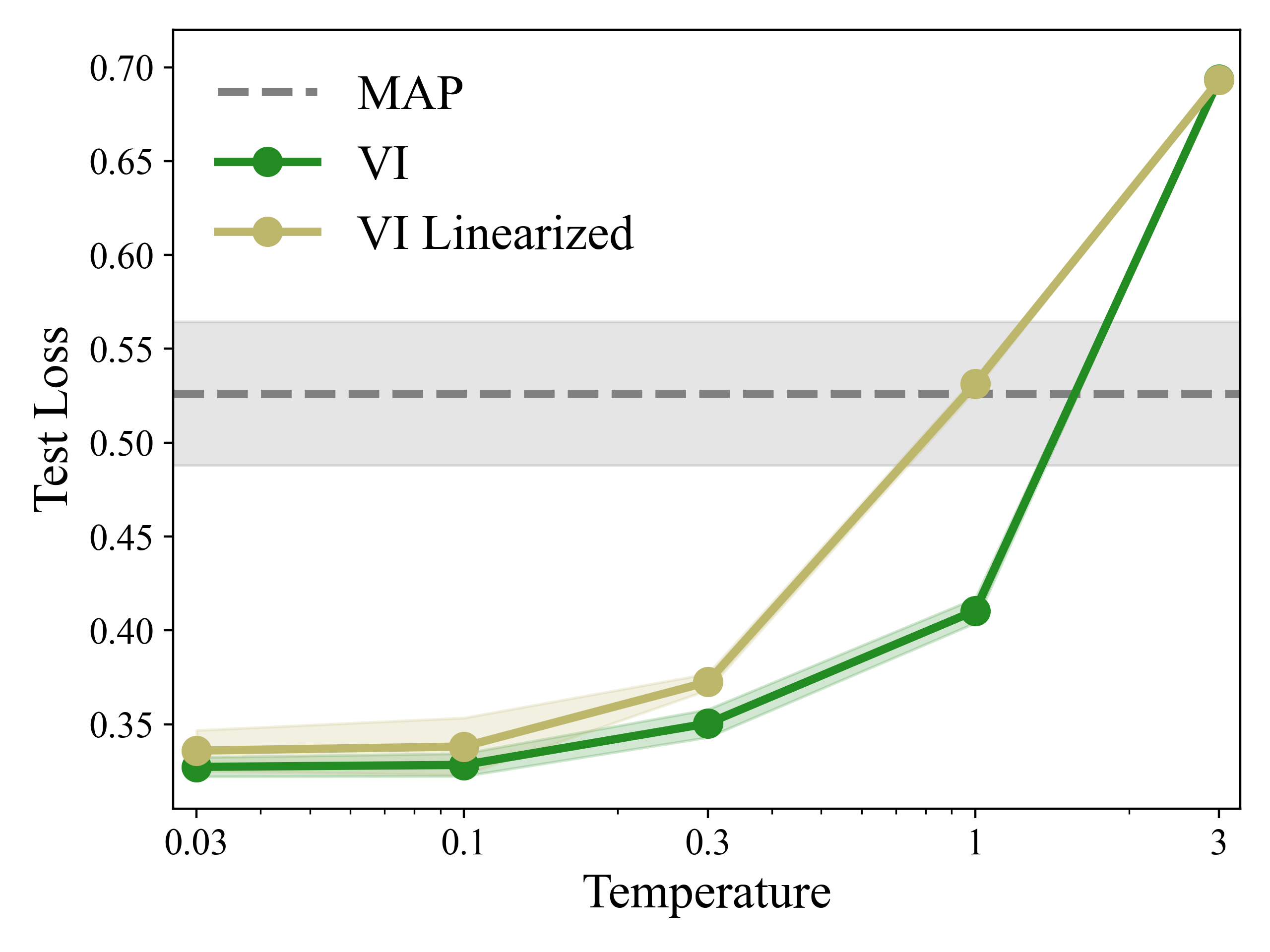}
    \end{subfigure}
    \hfill 
    \begin{subfigure}[b]{0.32\textwidth}
        \includegraphics[width=\textwidth]{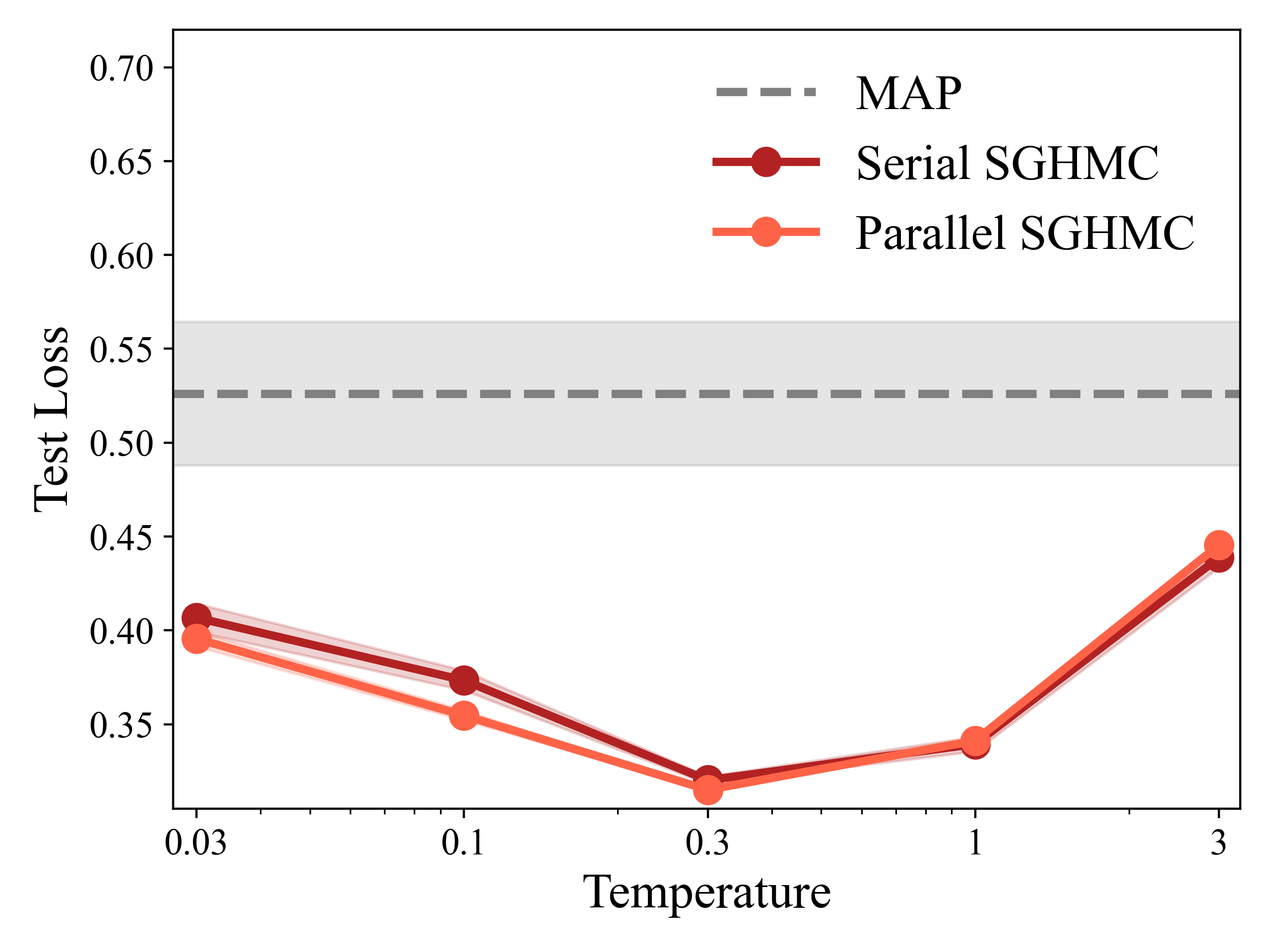}
    \end{subfigure}

    \caption{\textbf{Investigation into the cold-posterior effect} for a range of \posteriors{} algorithms for a CNN-LSTM model \citep{wenzel2020good} on the IMDB dataset \citep{maas2011learning}. Non-linearized Laplace EF and Laplace GGN are indistinguishable (left panel). All Gaussian approximations are diagonal. All approaches display error bars with one standard deviation over 5 random seeds.}
    \label{fig:cold_posterior}
\end{figure}

Following \cite{wenzel2020good, izmailov2021bayesian}, we train a CNN-LSTM model with 2.7 million parameters on the IMDB \citep{maas2011learning} for binary classification of positive or negative reviews. As in \cite{izmailov2021bayesian}, we use a diagonal Gaussian prior with variance $1/40$ for all methods.

The test set loss performance across a range of temperatures for Laplace, VI and SGMCMC approaches are depicted in Figure~\ref{fig:cold_posterior}, alongside the baseline optimization MAP approach. See Appendix~\ref{sec:details_cpe} for the same plots for accuracy and also optimization for the MLE (which overfits and performs significantly worse than MAP). There are several takeaways from Figure~\ref{fig:cold_posterior}, which we describe now.

The Bayesian methods, particularly VI and SGHMC can significantly outperform gradient descent for the MAP when evaluating on the test data. A simple Laplace approximation performs badly, however, improved performance is gained via the linearization discussed in Section~\ref{sec:bayes-learn}. Laplace GGN, which is equivalent to $F_C$ in this case (and in most machine learning models, see Appendix~\ref{subec:laplace}), improves over $F_E$, which is in agreement with \cite{kunstner2019limitations}. We observe that the reverse is true for VI and that the linearization is actually detrimental. We posit that this is due to the VI optimization acting in parameter space without knowledge of forward linearization. We observe a strong cold posterior effect for the Gaussian approaches with significantly superior performance achieved at $T<1$. However, for the (non-Gaussian) SGMCMC approaches, only a very mild cold posterior approach is observed. This is evidence in agreement with \cite{izmailov2021bayesian} that the cold posterior effect is more a result of the crudeness of posterior approximations than the of Bayesian learning itself \eqref{eq:bayes}. Finally, we observe that the parallel SGHMC approach (Bayesian deep ensemble) slightly outperforms the single trajectory serial approach, relatably Figure~\ref{fig:schematic}. And further, reducing the temperature deteriorates performance (observing from Section~\ref{sec:sg} and Appendix~\ref{sec:sgmcmc-opt} that traditional deep ensemble is recovered as parallel SGHMC with $T=0$).

\subsection{Continual learning with LoRA}\label{subsec:continual_lora}


The purpose of this experiment is to apply a simple Bayesian technique made accessible by \posteriors{}, namely, an empirical Fisher Laplace approximation, to mitigate catastrophic forgetting in an online setting with a large-scale model. Here, the composability of \posteriors{}, as well as the flexibility of the functional framework, is vital to make use of parameter-efficient fine-tuning (PEFT) \citep{peft} and the Llama 2 \citep{touvron2023llama} large language model from the PyTorch \citep{paszke2019pytorch} ecosystem. Extensive details can be found in Appendix~\ref{sec:details_continual_lora}.

\begin{figure}[ht]
    \centering
    \begin{minipage}[c]{0.48\textwidth}
        \centering
        \includegraphics[width=0.9\textwidth]{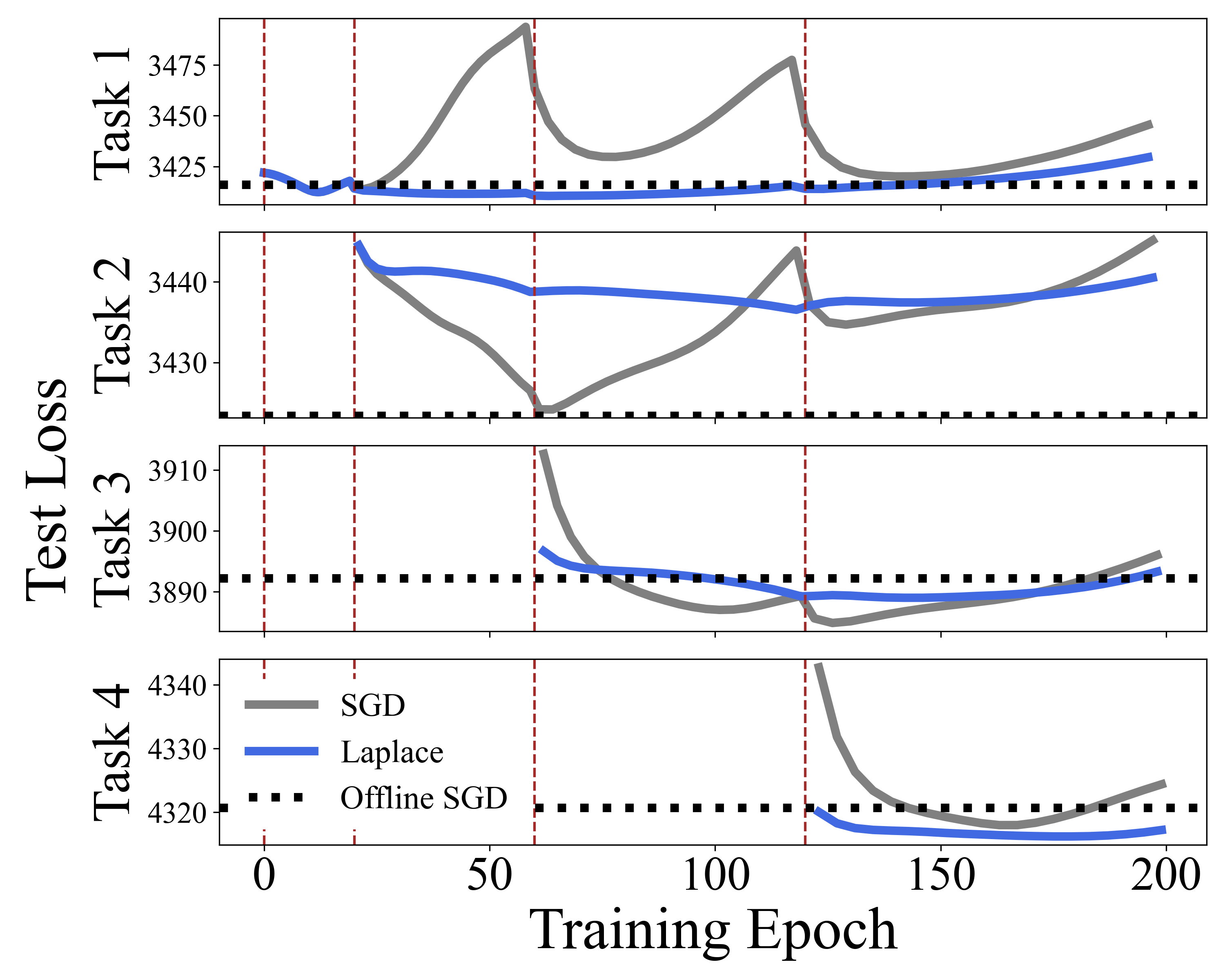}
        \caption{\textbf{Continual learning with Llama 2}. The online SGD and Laplace methods train one book after another, whilst the Offline SGD approach sees all books simultaneously, representing the network's capacity. Vertical dashed lines represent episode changes.}
        \label{fig:continual_a}
    \end{minipage}
    \hfill
    \begin{minipage}[c]{0.45\textwidth}
        \centering
        \includegraphics[width=0.99\textwidth]{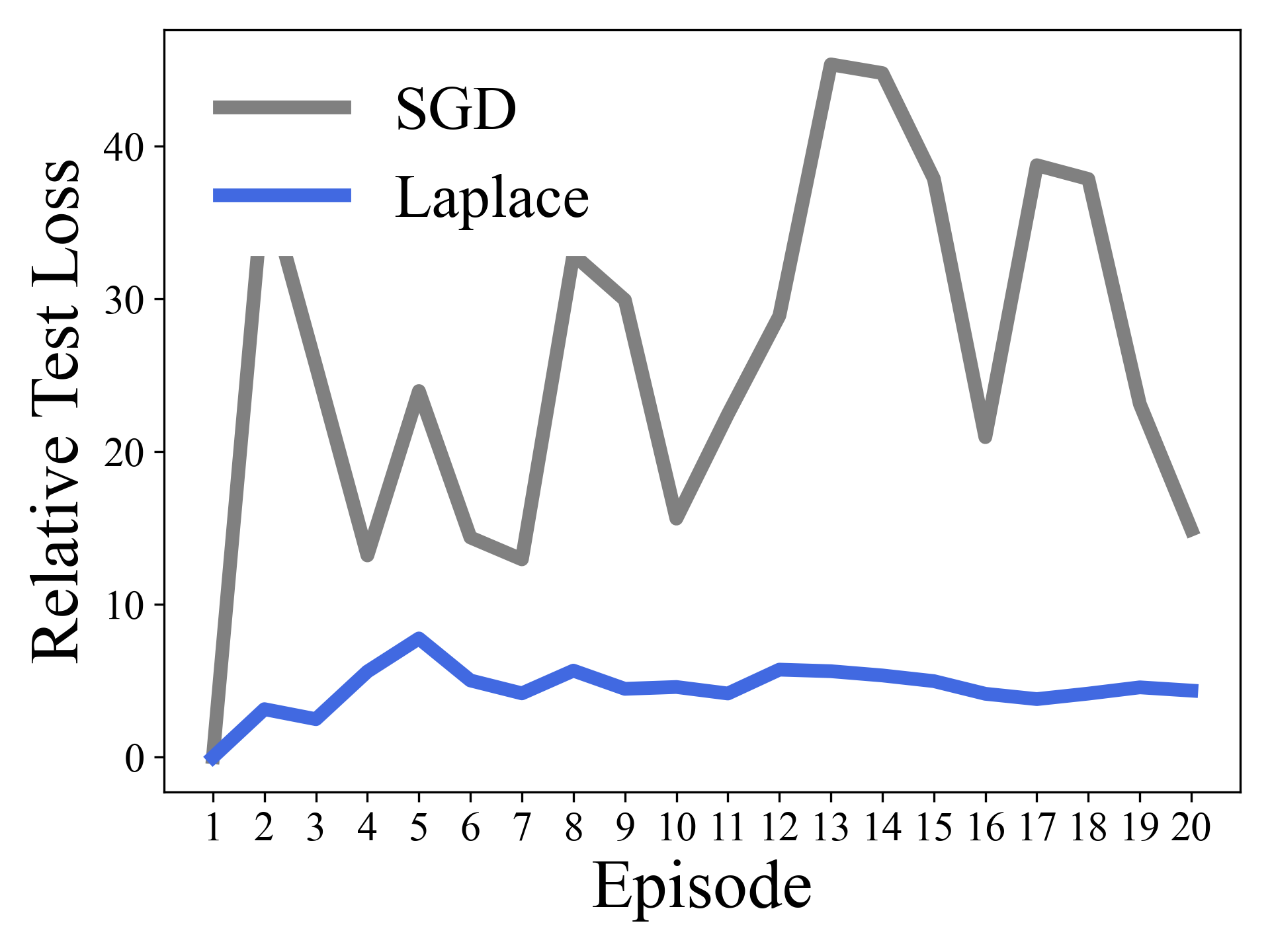}
        \caption{\textbf{Continual performance averaged over episodes seen thus far.} We use relative loss from a baseline (SGD trained on a single episode) for fairer averaging across episodes.}
        \label{fig:continual_b}
    \end{minipage}
\end{figure}

We consider the problem of episodal continual learning \citep{kirkpatrick2017overcoming, nguyen2017variational} where each episode's data, $y_n$, is one of 20 books from the PG19~\citep{raecompressive2019} dataset with 15\% of each book held out for testing. We compare simple online gradient descent with an approximation to the Bayesian online update \eqref{eq:bayes} $p(\theta \mid y_{1:n}) \propto p(\theta \mid y_{1:n-1}) p(y_n \mid \theta)$ in Figures~\ref{fig:continual_a} and \ref{fig:continual_b}.
Our baseline implements AdamW via \texttt{torchopt}'s integration within the posteriors{} unified API. The Laplace posterior from the previous episode is then used as the prior for the next episode following the suggested amendment to the EWC method \citep{kirkpatrick2017overcoming} recommended in \citep{huszar2018note}. We also train an offline version of the same model with access to all books to indicate the model's total capacity (although unrealistic of an online pipeline).

We use LoRA \citep{hu2021lora} to fine-tune the query, key and output weight matrices in the last attention layer of the 7B parameter model Llama 2 model \citep{touvron2023llama}. LoRA computes a low-rank approximation to the weight updates and is implemented in the PEFT library \cite{peft}, which integrates easily into functional \posteriors{} code. We use standard choices for the rank ($r=8$) and scaling ($\alpha=32$) parameters.

Figure~\ref{fig:continual_a} shows test loss for each episode, over all four episodes. The Laplace method maintains low loss in early tasks, while SGD forgets. For example, in the top row, the Laplace approximation stays low throughout training demonstrating that it continues to perform well on task 1 even though it is now being trained on data from tasks 2-4. In contrast, continually applying gradient descent quickly decreases the model's performance on task 1. Figure~\ref{fig:continual_b} shows the difference of test loss from baseline, averaged over all episodes seen thus far. We see that the Laplace approximation approach maintains low loss averaged over all tasks.

\subsection{Bayesian Llama 3}\label{subsec:bayesllama}

\begin{wrapfigure}[26]{r}{0.3\textwidth}
    \vspace{-60pt}
    \centering
    \includegraphics[width=0.28\textwidth]{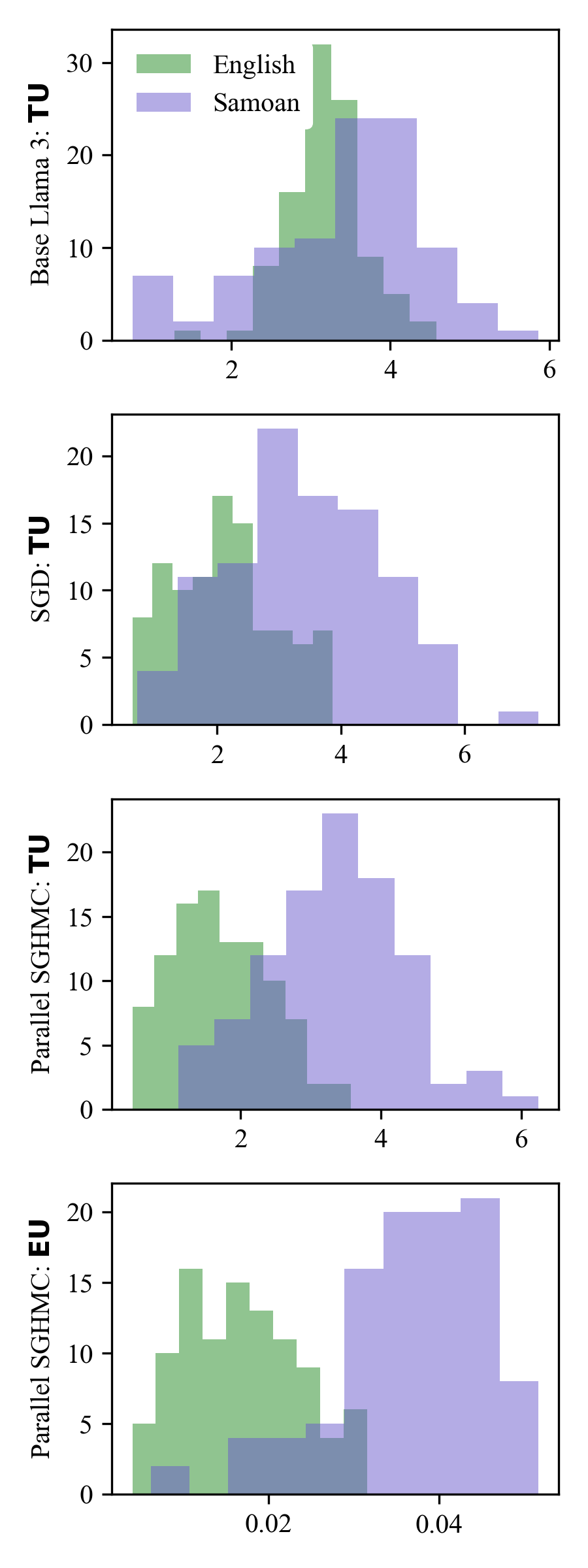}
    \caption{\textbf{Distributions of predictive uncertainties} with Llama 3.
    }
    \label{fig:bayesllama_hists}
\end{wrapfigure}


We now show how \posteriors{} can be used to fine-tune a very recent large language model to provide the capability to decompose uncertainty (\eqref{eq:uncs}) and provide improved out-of-distribution detection. The flexible and scalable design of \posteriors{} allows us to easily use subspace methods for fine-tuning a large pre-trained model in parallel using the novel Bayesian generalization of deep ensembles introduced in Section~\ref{subsec:psgmcmc}. Additionally, \posteriors{}' functional approach allows us to easily and efficiently map over the ensemble in forward calls at inference time. Extensive experimental details can be found in Appendix~\ref{sec:details_bayesllama}.

We fine-tune the last attention layer of the 8B Llama 3 model \citep{llama3modelcard} (resulting in 218 million trainable parameters) on the TQA \citep{kembhavi2017you} dataset, which consists of scientific textbooks. We fine-tune a baseline with SGD and a size 10 ensemble trained in parallel with \posteriors{}' SGHMC. We also compare against the base Llama 3 model without fine-tuning. We evaluate the models by asking them to complete 100 scientific statements (in-distribution) with the last word removed. We then repeat the experiment with the statements translated into Samoan, an out-of-distribution language for TQA \citep{kembhavi2017you}.

We can see from Figure~\ref{fig:bayesllama_hists} that although total uncertainty provides some indication for classifying the Samoan text as out-of-distribution, the epistemic uncertainty of parallel SGMHC offers a clearer split. Table~\ref{tab:auroc} validates this with the AUROC metric \citep{kuhn2023semantic}, whilst also displaying the categorical loss values for the next token of the English statements to verify a successful fine-tune. Overall, we can see a significant advantage in out-of-distribution detection via the Bayesian decomposition of uncertainty, whereas the point-estimate-based methods (MLE/MAP) cannot decompose uncertainty to differentiate epistemic uncertainty from aleatoric or semantic uncertainty which is key for open-ended natural language generation \citep{kuhn2023semantic}.

In Appendix~\ref{sec:details_bayesllama}, we include results for serial SGHMC, which interestingly fails to provide improved epistemic uncertainty quantification. We posit that this has to do with serial SGHMC staying isolated in a single mode of the posterior in this case, and therefore, averaging along the trajectory provides little over an SGD approach as in Figure~\ref{fig:schematic}.

\begin{table}
  \caption{\textbf{Out-of-distribution detection} with Bayesian Llama 3.}\label{tab:auroc}
  \centering
  \begin{tabular}{lcccc}
    \toprule
    & Base Llama 3 & SGD& \multicolumn{2}{c}{Parallel SGHMC}   \\
         & \bf{TU}    &  \bf{TU}  & \: \:  \bf{TU} & \bf{EU}  \\
    \midrule
    \textbf{AUROC} $\uparrow$ &   0.62 & 0.80 & \: \: 0.90 & \bf{0.95}  \\
    \textbf{Loss on English statements} $\downarrow$ &   4.68 & \textbf{4.20} & \multicolumn{2}{c}{4.40}  \\
    \bottomrule
  \end{tabular}
\end{table}

\section{Related work}\label{sec:related}

Section~\ref{sec:bayes-learn} (and more thoroughly Appendix~\ref{sec:survey}) summarizes much of the related work regarding Bayesian methodology. Additional comprehensive reviews can be found in \cite{wang2020survey} with \cite{fortuin2022priors} focussing on prior specifications (which we do not go into detail here but is handled with broad flexibility in \posteriors{}) and \cite{izmailov2021bayesian} providing extensive experiments with full-batch MCMC to give a closer indication of the behaviour of the true Bayesian posterior.


There is a broad selection of software for uncertainty quantification; however, to our knowledge and experience, none satisfy all five points in Section~\ref{sec:posteriors}. Stan \citep{carpenter2017stan} is widely used within the field of Bayesian statistics however does not support minibatching and is not easily extensible. Pyro \citep{bingham2019pyro} is a broad framework that supports a range of models and inference techniques; however, it is more heavyweight than \posteriors{} and harder to compose with existing PyTorch code. BlackJAX \citep{cabezas2024blackjax} (and optax \citep{deepmind2020jax}) heavily inspired the \posteriors{} functional approach. However, BlackJAX also has a primary focus on full-batch inference whilst \posteriors{} is minibatch-first and, like other JAX \citep{phan2019composable, detommaso2023fortuna, mocat2021} packages, does not compose with the PyTorch ecosystem.

\section{Conclusion and outlook}

We have presented a new software package \posteriors{} for scalable and extensible uncertainty quantification with PyTorch \citep{paszke2019pytorch} models. Through numerical experiments, we have demonstrated how \posteriors{} can be used to achieve a range of benefits over optimization-based approaches and composing with PyTorch toolkits. In particular, we highlighted that the cold posterior effect mostly occurs in Gaussian approximations to the Bayesian posterior as well as demonstrating the benefits of Bayesian learning in the very practical and large-scale setting of free-form text generation with large language models.

We are excited by the future prospects for \posteriors{}. Its functional API makes it easy to maintain and extend; we already have a long list of methods and approaches to add. \posteriors{} is fully open source, and we are keen to foster a community for feedback for feature requests and encourage contributions. The framework from Section~\ref{sec:sg} allows for simultaneous research into optimization and sampling methods. Combining that with \posteriors{}' broad range of utility functions (that, to our knowledge, are not available elsewhere in the PyTorch ecosystem), means that implementation and research into second-order techniques \citep{martens2010deep, martens2020new} and improved discretization methods \citep{leimkuhler2018ensemble} represent accessible avenues for future work in both optimization and SGMCMC.


\bibliographystyle{abbrvnat}
\bibliography{references}

\appendix

\section{Survey of scalable Bayesian learning}\label{sec:survey}

We now provide a more thorough review of methods for scalable Bayesian learning, extending Section~\ref{sec:bayes-learn}.

\subsection{Laplace approximation}\label{subec:laplace}

As discussed, the Laplace approximation \citep{mackay1992practical,daxberger2021laplace} forms a Gaussian approximation to the posterior distribution $p(\theta \mid y_{1:N}) \approx \gauss (\theta \mid \theta_\text{MAP}, \hat{\Sigma})$ based on the second order Taylor expansion around the posterior mode $\theta_\text{MAP} = \argmax_\theta p(\theta \mid y_{1:N})$:
\begin{align*}
\log p(\theta \mid y_{1:N}) &\approx (\theta - \theta_\text{MAP})^\top  \nabla^2_\theta \log p(\theta_\text{MAP} \mid y_{1:N}) (\theta - \theta_\text{MAP}) +\text{const} ,\\
&\approx \log \gauss (\theta \mid \theta_\text{MAP}, \hat{\Sigma}) +\text{const}.
\end{align*}
where $\nabla_\theta \log p(\theta \mid y_{1:N}) = 0$ at $\theta = \theta_\text{MAP}$.

The Taylor expansion implies the use of the negative inverse Hessian $-\nabla^2_\theta \log p(\theta_\text{MAP} \mid y_{1:N})^{-1}$ as the covariance $\hat{\Sigma}$, however this is not guaranteed to be positive semidefinite and is therefore invalid as a covariance matrix. Instead, we can draw from the Bernstein-von Mises theorem \citep{hartigan1983asymptotic}:
$$
p(\theta \mid y_{1:N}) \underset{N\to \infty}{\to} \gauss\left(\theta \mid \theta_\text{MAP}, N^{-1} \F(\theta_\text{MAP})^{-1}\right),
$$
with Fisher information matrix (which is guaranteed to be positive semidefinite):
\begin{align}
    \F(\theta) = \E_{p(y\mid \theta)}[\nabla^2_\theta \log p(\theta \mid y)] = \E_{p(y\mid \theta)}[\nabla_\theta\log p(\theta \mid y) \nabla_\theta\log p(\theta \mid y)^\top].
\end{align}
In the supervised learning setting with inputs $x$, the Fisher information matrix becomes $\F(\theta) = \E_{p(x, y\mid \theta)}[\nabla^2_\theta \log p(\theta \mid x, y)]$ which is intractable as we do not have access to the distribution over inputs $p(x)$. Two popular approximations use the training data $\hat{p}(x, y) = N^{-1}  \sum_{i=1}^N \delta(x, y \mid x^i, y^i)$. The \textit{empirical Fisher}:
\begin{equation*}
\F_\text{E}(\theta) = \E_{\hat{p}(x, y)} [\nabla_\theta\log p(\theta \mid y) \nabla_\theta\log p(\theta \mid y)^\top],
\end{equation*}
and \textit{conditional Fisher}:
\begin{equation}\label{eq:cond_fish}
\F_\text{C}(\theta) = \E_{\hat{p}(x) p(y \mid x, \theta)} [\nabla_\theta\log p(\theta \mid y) \nabla_\theta\log p(\theta \mid y)^\top],
\end{equation}
with the conditional Fisher generally recommended \citep{kunstner2019limitations}.

The integral with respect to $p(y \mid x, \theta)$ in \eqref{eq:cond_fish} may initially seem difficult to compute. Fortunately, things are simplified in the common machine learning setting where the likelihood has the form $p(y \mid x, \theta) = p(y \mid f_\theta(x))$ where $p(y \mid z)$ is the probability density function for an exponential family distribution with natural parameter $z$. That is
\begin{equation*}
    \log p(y\mid z) = z^\top T(y) - \log \int z^\top T(y) dy,
\end{equation*}
for some $T(y)$. This setting includes the common cases of classification (softmax with cross-entropy loss) and regression (mean squared error loss). Then we have an exact equivalence between $F_C(\theta)$ and the Generalized Gauss-Newton matrix (GGN, \cite{martens2020new}) matrix:
\begin{equation*}
    G(\theta) = \mathbb{E}_{\hat{p}(x, y)}[\nabla_\theta f_\theta(x)^\top
    H_L(x, y)
    \nabla_\theta f_\theta(x)],
\end{equation*}
where $\nabla_\theta f_\theta(x)$ is the Jacobian of the forward function and
\begin{equation*}
    H_L(x, y) = [- \nabla^2_z p(y \mid z)]_{z = f_\theta(x)},
\end{equation*}
is the Hessian of the loss function evaluated at $z = f_\theta(x)$.

\subsubsection*{Caveats and computational bottlenecks}

Laplace approximation training consists of first determining the posterior mode $\theta_\text{MAP}$ through gradient descent. This is followed by a single additional epoch to calculate the approximate covariance. It is, thus, cheap to compute in the large data regime. However, in the large parameter regime, storing the covariance matrix with $\bigOh(d^2)$ elements becomes a memory bottleneck as well as the required inverting and sampling operations, which typically have runtime $\bigOh(d^3)$. As such, approximations are used, such as diagonal \citep{mackay1992practical}, K-FAC \citep{ritter2018scalable} and low-rank \citep{daxberger2021laplace} with reduced $\bigOh(d)$ memory. The main caveat for Laplace approximations is the accuracy of a Gaussian approximation under the extreme non-linearities in a neural network posterior \citep{izmailov2021bayesian}, which applies even further to reduced memory approximations.

\subsection{Variational inference}\label{subsec:vi}

Variational inference seeks a parameterized distribution $q_\phi(\theta)$ that minimizes some distance from the target posterior distribution $p(\theta \mid y_{1:N})$. Most commonly \citep{blei2017variational} we assume a Gaussian variational distribution $q_\phi(\theta) = \gauss (\theta \mid \mu, \Sigma)$ with $\phi = (\mu, \Sigma)$ and the distance to be minimized is the KL divergence:
\begin{align}
    0 \leq \text{KL}[q_\phi(\theta) \mid \mid p(\theta \mid y_{1:N})] =
    \log p(y_{1:N}) + \underbrace{\E_{q_\phi(\theta)}\left[\log \frac{q_\phi(\theta)}{p(\theta)} - \log p(y_{1:N} \mid \theta)  \right]}_{\text{NELBO}(\phi)}.
\end{align}

Thus, the problem becomes one of optimization to minimize the \textit{negative evidence lower bound} objective $\text{NELBO}(\phi)$, which can be achieved with stochastic gradient descent alongside the reparameterization trick \citep{rezende2014stochastic} to obtain unbiased gradients of the expectation.

\subsubsection*{Caveats and computational bottlenecks}

Notably, the $\text{NELBO}(\phi)$ objective can be estimated stochastically by sampling from $q_\phi(\theta)$ and even minibatching the data $y_{1:N}$. Therefore, training is scalable in the large data regime. The caveat regarding the crudeness of a Gaussian approximation to the posterior also applies here, except that variational inference minimizes a formal divergence from the posterior rather than relying on Bernstein-von Mises asymptotics. However, in the large parameter regime, the $\text{NELBO}(\phi)$ objective requires high dimensional integration at every optimization step, which can suffer from high variance even with the available tricks \citep{blei2017variational}.

\subsection{Stochastic gradient Markov chain Monte Carlo}\label{subsec:SGMCMC}

A completely different paradigm to posterior approximation is to construct a Monte Carlo approximation $p(\theta \mid y_{1:N}) \approx N^{-1}\sum_{k=1}^K \delta(\theta \mid \theta_k)$. This is most commonly achieved by evolving a stochastic differential equation (SDE) and collecting samples along the trajectory. Following \cite{ma2015complete}, an SDE has $\pi(z)$ as a stationary distribution (and therefore provides an asymptotically unbiased Monte Carlo approximation) if and only if it is of the form
\begin{align}\label{eq:sde}
    dz = [\D(z) + \Q(z)]\nabla \log \pi(z) dt + \nabla \cdot [\D(z) + \Q(z)] dt + \sqrt{2 \D(z)} dw. 
\end{align}
with symmetric positive-semidefinite $\D(z)$ and skew symmetric $\Q(z) = -\Q(z)^\top$. Here $z = (\theta, \omega)$ where $\omega$ are any auxiliary parameters such as momenta and thermostat \citep{ma2015complete, ding2014bayesian}. The distribution $\pi(z)$ is thus an extended target with $p(\theta \mid y_{1:N})$ marginal in $\theta$. The matrices $\D(z)$ and $\Q(z)$ are often chosen as independent of $z$, meaning the matrix divergence term vanishes.

The exact simulation of such a non-linear SDE is intractable, and as such, approximate discretization schemes such as the Euler-Maruyama \citep{welling2011bayesian} method are employed. Traditional full-batch Markov chain Monte Carlo (MCMC) schemes \citep{andrieu2003introduction} use a Metropolis-Hastings step to ensure convergence of the Monte Carlo approximation is preserved through discretization. In the minibatch setting, the Metropolis-Hastings ratio cannot be estimated easily. Instead, we can use an unbiased stochastic gradient $\widehat{\nabla \log \pi}(z) \sim \gauss( \cdot \mid \nabla \log \pi(z), \mat{B}(z))$, then  as noted in \cite{welling2011bayesian} the stochasticity arising from minibatching, $\mathrm{B}(z)$, decreases like $\bigOh(\epsilon_t^2)$ for learning rate $\epsilon_t$. A Robbins-Monro \citep{robbins1951stochastic} learning rate schedule $\sum \epsilon_t^2 < \sum \epsilon_t = \infty$ ensures an asymptotically unbiased Monte Carlo approximation \citep{welling2011bayesian}. Additional enhancements can be made if an approximation to $\mat{B}(z)$ is available \citep{ma2015complete}.

We detail specific implementations of SGMCMC in Section~\ref{sec:sgmcmc}.

\subsubsection*{Caveats and computational bottlenecks}

The stochastic gradient formulation of MCMC permits easy application to the minibatch and large data regime. However, the decreasing learning rate requirement results in diffusive behaviour \citep{alexos2022structured, li2016preconditioned, bieringer2023adammcmc} where the discretized SDE puts excessive weight on the noise term $dw$ over the informative gradient term. This is problematic since exploration is very much desired for MCMC. Often, a constant learning rate is used in practice \citep{leimkuhler2018ensemble} with an asymptotic bias incurred. In the large parameter regime, the formulation of a Monte Carlo approximation becomes a memory bottleneck. Typical lower dimensional MCMC procedures take many thousands of samples, but due to memory constraints, only a significantly smaller sample size is available for large neural network parameters, resulting in a crude approximation. A notable further caveat is that, in contrast to the Gaussian approximations, the resulting Monte Carlo approximation does not provide access to a density with pointwise evaluations and cannot be used as a prior for subsequent Bayesian updates in an online regime \eqref{eq:bayes}.

\subsection{Notable mentions}\label{subsec:notables}

Perhaps the most common route to uncertainty quantification in neural networks is dropout \citep{gal2016dropout, gal2017concrete}, where model parameters or neurons are randomly removed from each forward call. This approach can be effective and has a Bayesian interpretation in the posterior probability space of inclusion/exclusion, but this is naturally more restricted than the posterior approximations above.

Stochastic weight averaging (SWA, \cite{izmailov2018averaging}) and SWA-Gaussian (SWAG, \cite{maddox2019simple}) are similar in spirit to SGMCMC as they use samples from a single trajectory to generate a point or Gaussian approximation to the posterior respectively. Another option for formulating a Gaussian distribution over parameters is to consider each minibatch as a small-scale Bayesian update; local linearization of this update is the paradigm of the extended Kalman filter \citep{duran2022efficient}, which also has natural gradient descent \citep{ollivier2018online} and variational inference interpretations \citep{zhang2018noisy}. These methods are easy to implement but again suffer from the caveats of Gaussian posterior approximations above.

\section{SGMCMC algorithms}\label{sec:sgmcmc}

Recall the generic SDE \cite{ma2015complete} from Section~\ref{sec:sg} for targeting the tempered $\pi_{\T}(z) \propto \pi(z)^{\frac{1}{N\T}}$ where we set $\T = N^{-1}$ to target the Bayesian posterior
\begin{align*}
    dz &= [\D(z) + \Q(z)] N^{-1} \nabla\log \pi(z) dt + \T \nabla \cdot [\D(z) + \Q(z)] dt + \sqrt{2 \T \D(z)} dw,
\end{align*}
for symmetric positive-semidefinite $\D(z)$ and skew symmetric $\Q(z) = -\Q(z)^\top$. With $z = (\theta, \omega)$ for any auxiliary variables $\omega$ and $\int \pi(\theta, \omega) d \omega = \pi(\theta) = p(\theta \mid y_{1:N})$.

The matrix divergence $\nabla \cdot \mat{M}(z) \in \real^d$ is defined \cite{ma2015complete} as 
\begin{equation*}
[\nabla \cdot \mat{M}(z)]_i = \sum_{j=1}^d \frac{\partial}{\partial z_j} [\mat{M}(z)]_{ij},
\end{equation*}
and vanishes in the common case that the preconditioners are set to be independent of $z$, i.e. $\D(z) = D$  and $\Q(z) = Q$.

We now cover the most popular SGMCMC algorithms and those implemented in \posteriors{}, all of which are also found in \cite{ma2015complete}.

\subsection{SGLD}\label{subsec:sgld}
The simplest SGMCMC algorithm is stochastic gradient Langevin dynamics \citep{welling2011bayesian}, which contains no auxiliary variables (i.e. $z=\theta$) and is represented by $\D = \mathbb{I}$ and $\Q = 0$. This gives continuous-time SGLD:
\begin{equation*}
    d\theta = N^{-1} \nabla \log \pi(\theta) dt + \sqrt{2 \T}dw.
\end{equation*}
And Euler-Maruyama discretization:
\begin{equation*}
    \theta_{k+1} = \theta_k + \epsilon N^{-1} \nabla \log \pi(\theta_k) + \sqrt{2 \epsilon \T} \zeta_k, \tag*{$\zeta_k \sim \gauss (\zeta \mid 0, \mathbb{I}).$}
\end{equation*}

\subsection{SGHMC}\label{subsec:sghmc}
Stochastic gradient Hamiltonian Monte Carlo \cite{chen2014stochastic} extends $z = (\theta, m)$ to include momenta $m$ with Gaussian target distribution
\begin{equation*}
    \log \pi(\theta, m) = \log \pi (\theta) - \frac{1}{2\sigma^2} m^\top m + \text{const}.
\end{equation*}
Choosing
$
\D = \begin{pmatrix}
    0 & 0 \\
    0 & \alpha \mathbb{I}
\end{pmatrix}
$
and 
$
\Q = \begin{pmatrix}
    0 & -\mathbb{I} \\
    \mathbb{I} & 0
\end{pmatrix}
$
gives continuous-time SGHMC
\begin{align*}
    d\theta &= \sigma^{-2} m dt, \\
    dm &= N^{-1} \nabla \log \pi(\theta) dt - \alpha \sigma^{-2} m dt + \sqrt{2 \T \alpha }dw,
\end{align*}
and Euler-Maruyama discretization:
\begin{align*}
    \theta_{k+1} &= \theta_k + \epsilon \sigma^{-2} m_k, \\
    m_{k+1} &= m_k + \epsilon N^{-1} \nabla \log \pi(\theta) - \epsilon \sigma^{-2} \alpha m_k + \sqrt{2 \epsilon \T\alpha} \zeta_k, \tag*{$\zeta_k \sim \gauss (\zeta \mid 0, \mathbb{I}).$}
\end{align*}

\subsection{SGNHT}\label{subsec:sgnht}
The stochastic gradient Nosé-Hoover thermostat (SGNHT) algorithm \citep{ding2014bayesian} uses ideas from thermodynamics \citep{leimkuhler2018ensemble} to introduce an additional scalar variable $\xi$ that automates the selection of the $\alpha$ friction parameter in SGHMC. We have $z = (\theta, m, \xi)$ targeting
\footnote{Note that the SGNHT target distribution for $\xi$ is erroneously claimed as $\frac{1}{2d}(\xi - \alpha)^2$ in \cite{ma2015complete} and \cite{nemeth2021stochastic}}
\begin{equation*}
    \log \pi(\theta, m, \xi) = \log \pi (\theta) - \frac{1}{2\sigma^2} m^\top m - 
    \frac{d}{2}(\xi - \alpha)^2 + 
    \text{const}.
\end{equation*}
Choosing
\begin{equation*}
\D(\theta, m, \xi) = \begin{pmatrix}
    0 & 0 & 0 \\
    0 & \alpha \mathbb{I} & 0 \\
    0 & 0 & 0
\end{pmatrix},
\quad
\Q(\theta, m, \xi) = \begin{pmatrix}
    0 & -\mathbb{I} & 0 \\
    \mathbb{I} & 0 & m/d \\
    0 & -m^\top/d & 0 
\end{pmatrix},
\end{equation*}
gives continuous-time SGNHT
\begin{align*}
    d\theta &= \sigma^{-2} m dt, \\
    dm &= N^{-1} \nabla \log \pi(\theta) dt - \xi \sigma^{-2} m dt + \sqrt{2\T \alpha}dw, \\
    d\xi &= (\sigma^{-2} d^{-1} m^\top m - \T) dt,
\end{align*}
and Euler-Maruyama discretization:
\begin{align*}
    \theta_{k+1} &= \theta_k + \epsilon \sigma^{-1} m_k, \\
    m_{k+1} &= m_k + \epsilon N^{-1} \nabla \log \pi(\theta) - \epsilon \sigma^{-2} \xi_k m_k + \sqrt{2 \epsilon \T \alpha} \zeta_k, \tag*{$\zeta_k \sim \gauss (\zeta \mid 0, \mathbb{I}),$} \\
    \xi_{k+1} &= \xi_k + \epsilon \left(\sigma^{-2} d^{-1} m_k^\top m_k - \T\right).
\end{align*}
Although SGNHT has the same tuning parameters $(\epsilon, \sigma, \alpha)$ as SGHMC, we note the $\alpha$ parameter only controls the noise level for the $m$ update and is absent from the deterministic dynamics, with even $\alpha = 0$ representing a reasonable choice.

\section{SGMCMC for optimization}\label{sec:sgmcmc-opt}

Let us consider gradient descent (with momenta) \citep{sutskever2013importance}, parameterization from PyTorch, for minimising $U(\theta)$:
\begin{align*}
    \theta_{k+1} &= \theta_k + \gamma m_k, \\
    m_{k+1} &= \mu m_k - (1 - \tau) \nabla U(\theta_k).
\end{align*}
Here vanilla SGD $\theta_{k+1} = \theta_k - \gamma \nabla U(\theta_k)$ is obtained with $\mu = 0$ and $\tau = 0$.

And also SGHMC \ref{subsec:sghmc} with $\T=0$ for $U(\theta) = - N^{-1} \log \pi(\theta) = -N^{-1} \log p(\theta \mid y_{1:N})$:
\begin{align*}
    \theta_{k+1} &= \theta_k + \epsilon \sigma^{-2} m_k, \\
    m_{k+1} &= (1 - \epsilon \sigma^{-2} \alpha) m_k - \epsilon \nabla U(\theta_k).
\end{align*}

We see an equivalence between SGD and SGHMC ($T=0$) \citep{wenzel2020good} with
\begin{equation}\label{eq:sghmc_reparam}
\begin{array}{rcl}
\gamma & = & \epsilon \sigma^{-2}, \\
\mu & = & 1 - \epsilon \sigma^{-2} \alpha, \\
\tau & = & 1 - \epsilon,
\end{array}
\quad \iff \quad
\begin{array}{rcl}
\epsilon & = & 1 - \tau, \\
\sigma^{-2} & = & \gamma (1-\tau)^{-1}, \\
\alpha & = & (1-\mu)\gamma^{-1}.
\end{array}
\end{equation}

We note that the transition from SGMCMC to optimization and back is not ubiquitous amongst methods. Indeed, the Euler-Maruyama SGNHT sampling algorithm \ref{subsec:sgnht}, does not result in a functional optimization algorithm at the zero temperature limit as the value thermostat can only increase and results in divergant dynamics. Also, the popular Adam \citep{kingma2014adam} optimizer does not fit the general sampling SDE \eqref{eq:sde} due to its use of elementwise operations, which do not have a convenient matrix-vector-product form. \posteriors{}' framework makes these mathematical considerations transparent and an easily accessible avenue for research.

\section{Experiment details: Cold posterior effect}\label{sec:details_cpe}

We now give additional details for the experiment into generalization capabilities of the Bayesian methods in \posteriors{} and the cold posterior effect, from Section~\ref{subsec:cpe}. Fully reproducible code can be found at \href{https://github.com/normal-computing/posteriors/blob/main/examples/imdb}{\texttt{posteriors/examples/imdb}}.

\begin{figure}[ht]
    \centering
    \begin{subfigure}[b]{0.32\textwidth}
        \includegraphics[width=\textwidth]{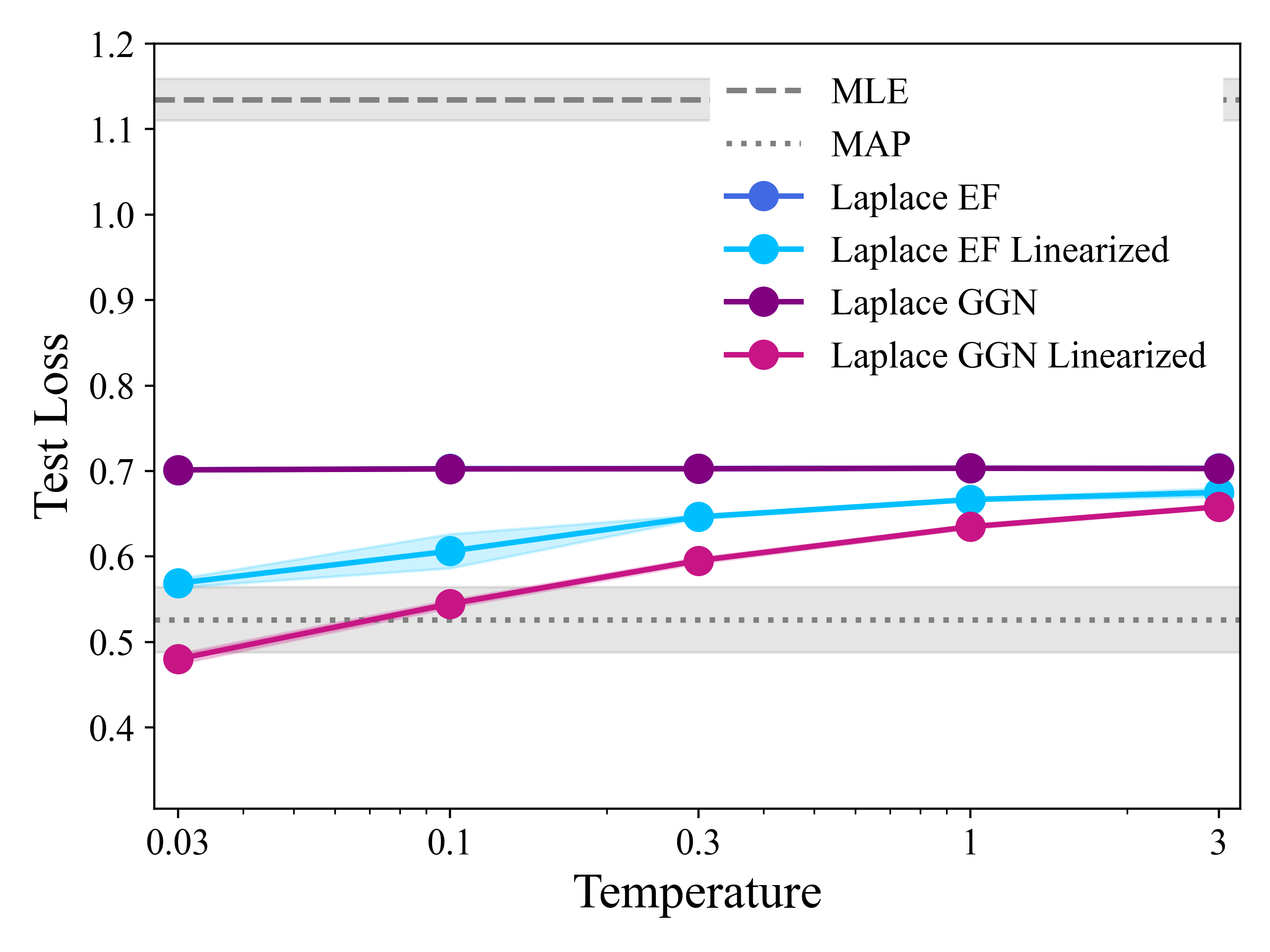}
    \end{subfigure}
    \hfill 
    \begin{subfigure}[b]{0.32\textwidth}
        \includegraphics[width=\textwidth]{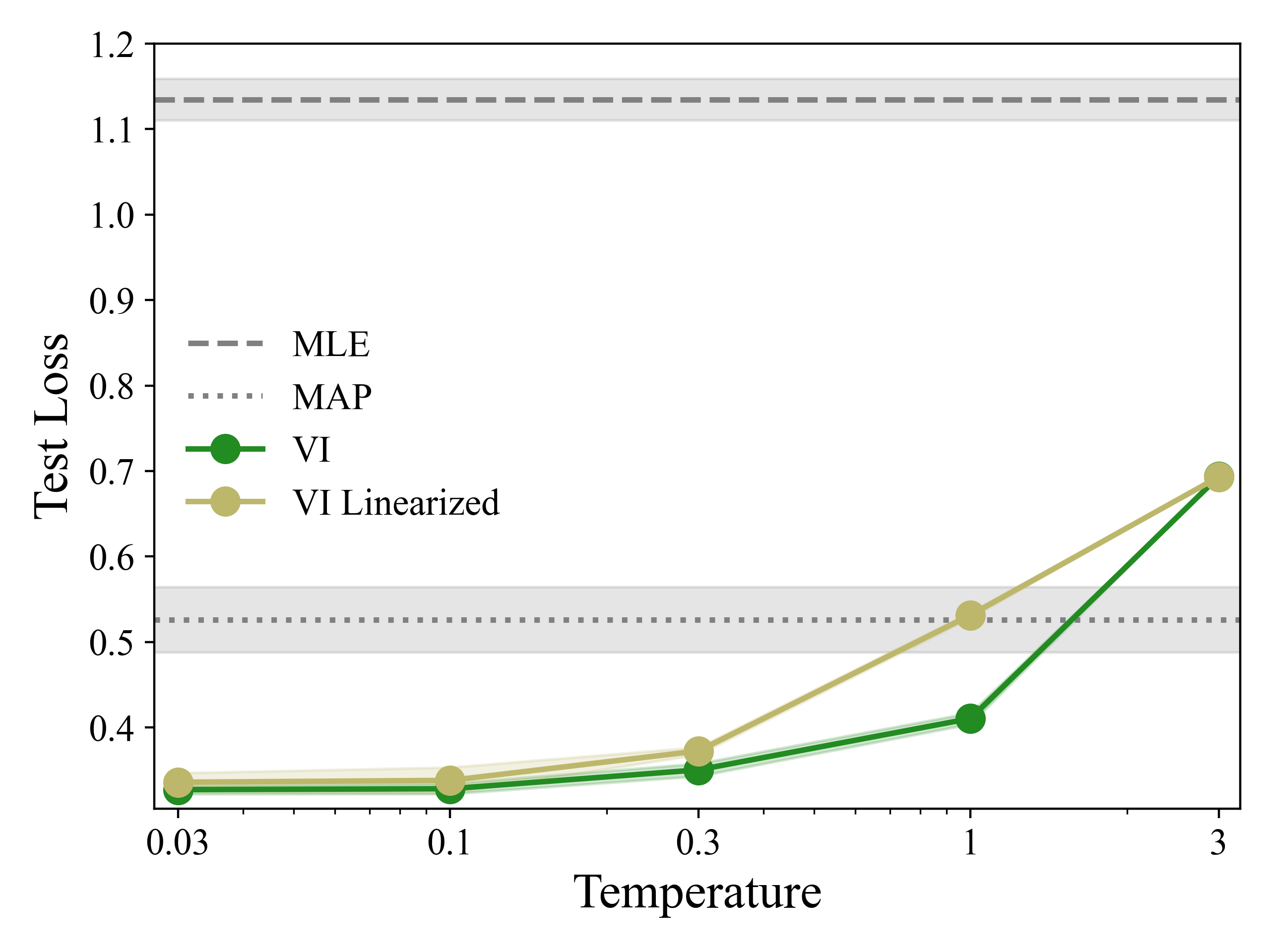}
    \end{subfigure}
    \hfill 
    \begin{subfigure}[b]{0.32\textwidth}
        \includegraphics[width=\textwidth]{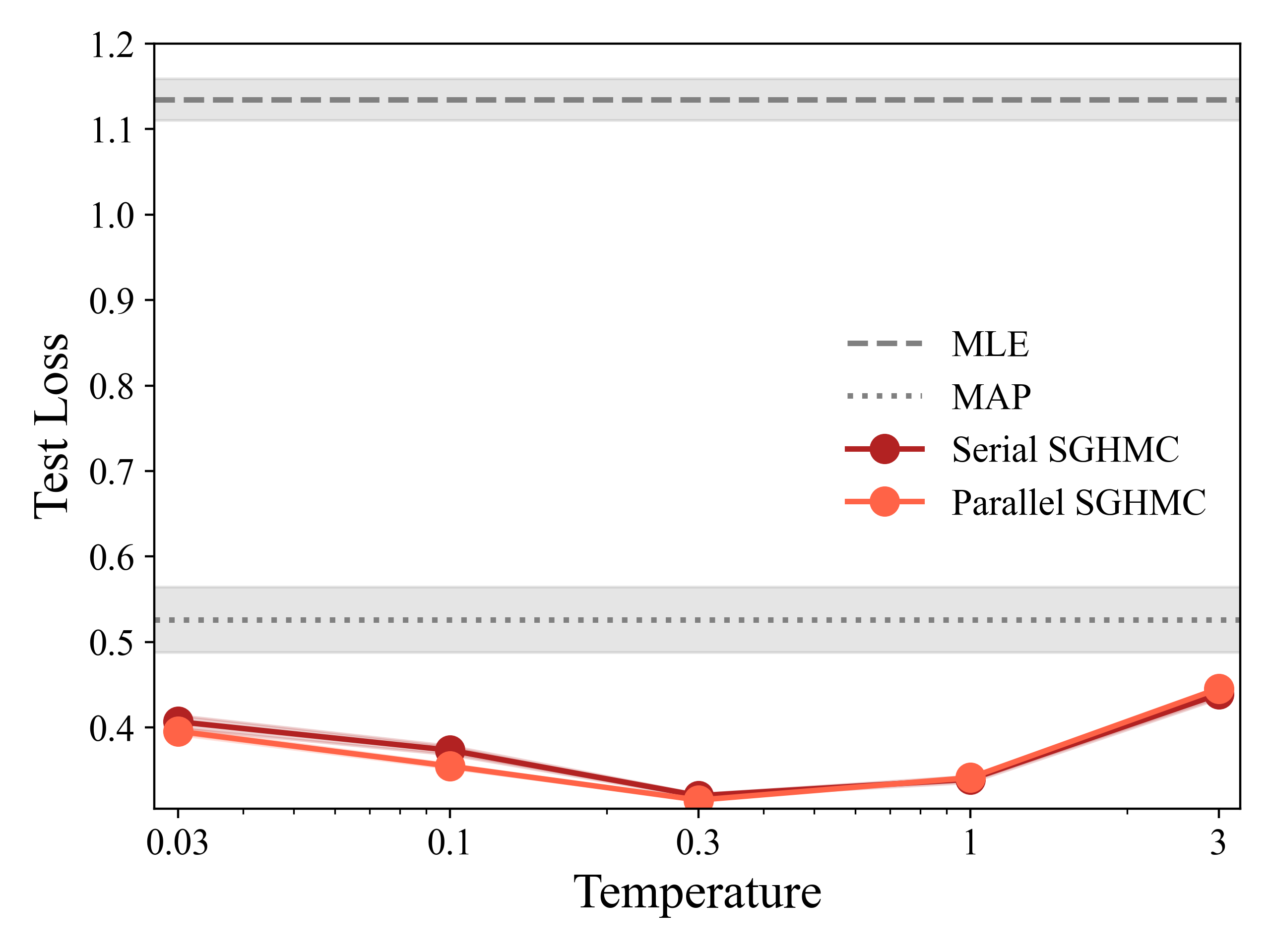}
    \end{subfigure}

    \caption{\textbf{Cold posterior test losses} (lower is better) for a range of \posteriors{} algorithms and temperatures for the CNN-LSTM model \citep{wenzel2020good} on the IMDB dataset \citep{maas2011learning} from Section~\ref{subsec:cpe} (same as Figure~\ref{fig:cold_posterior} with MLE added). Laplace EF is indistinguishable from Laplace GGN.}
    \label{fig:cold_posterior_with_mle}
\end{figure}

\begin{figure}[ht]
    \centering
    \begin{subfigure}[b]{0.32\textwidth}
        \includegraphics[width=\textwidth]{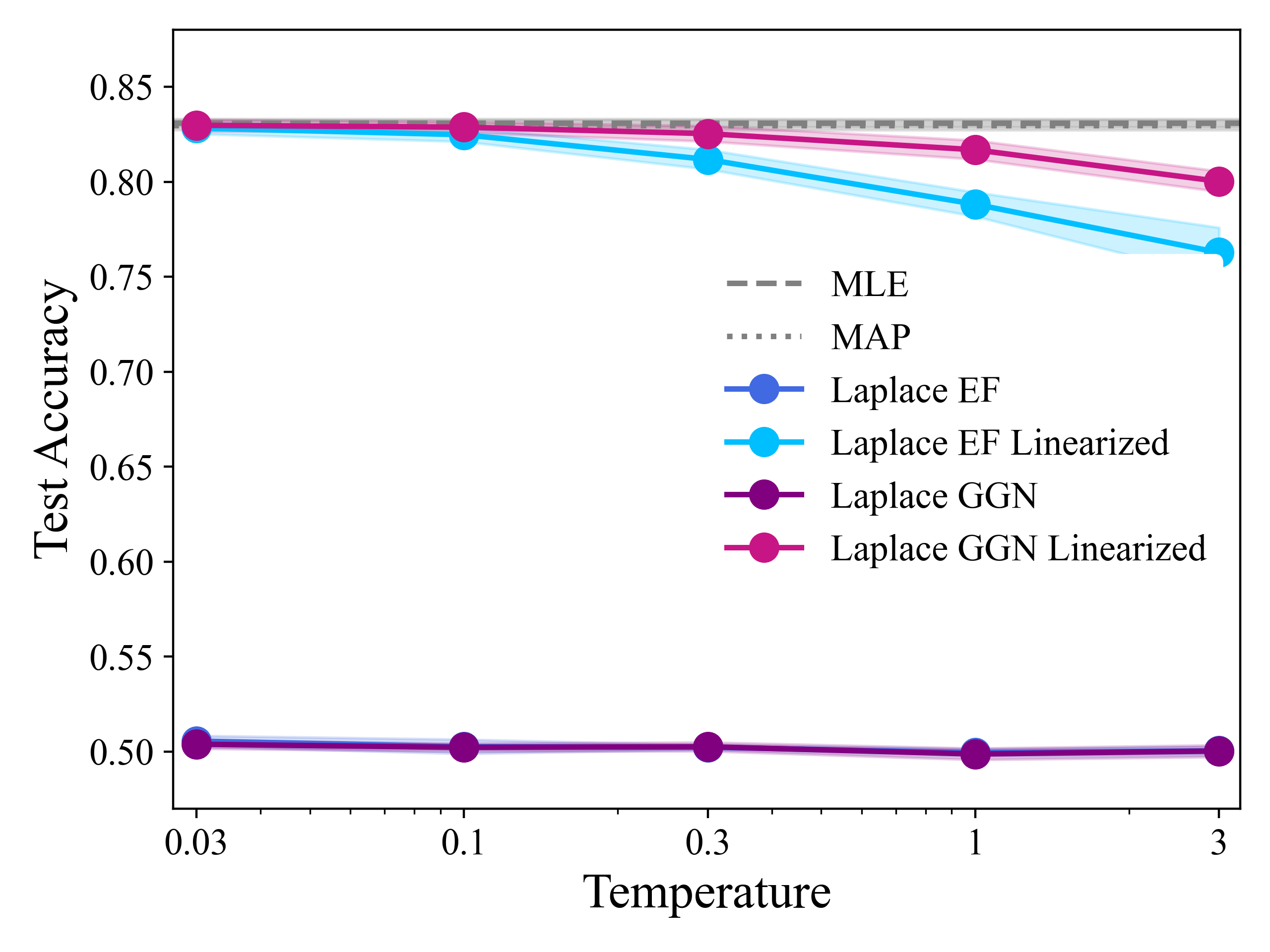}
    \end{subfigure}
    \hfill 
    \begin{subfigure}[b]{0.32\textwidth}
        \includegraphics[width=\textwidth]{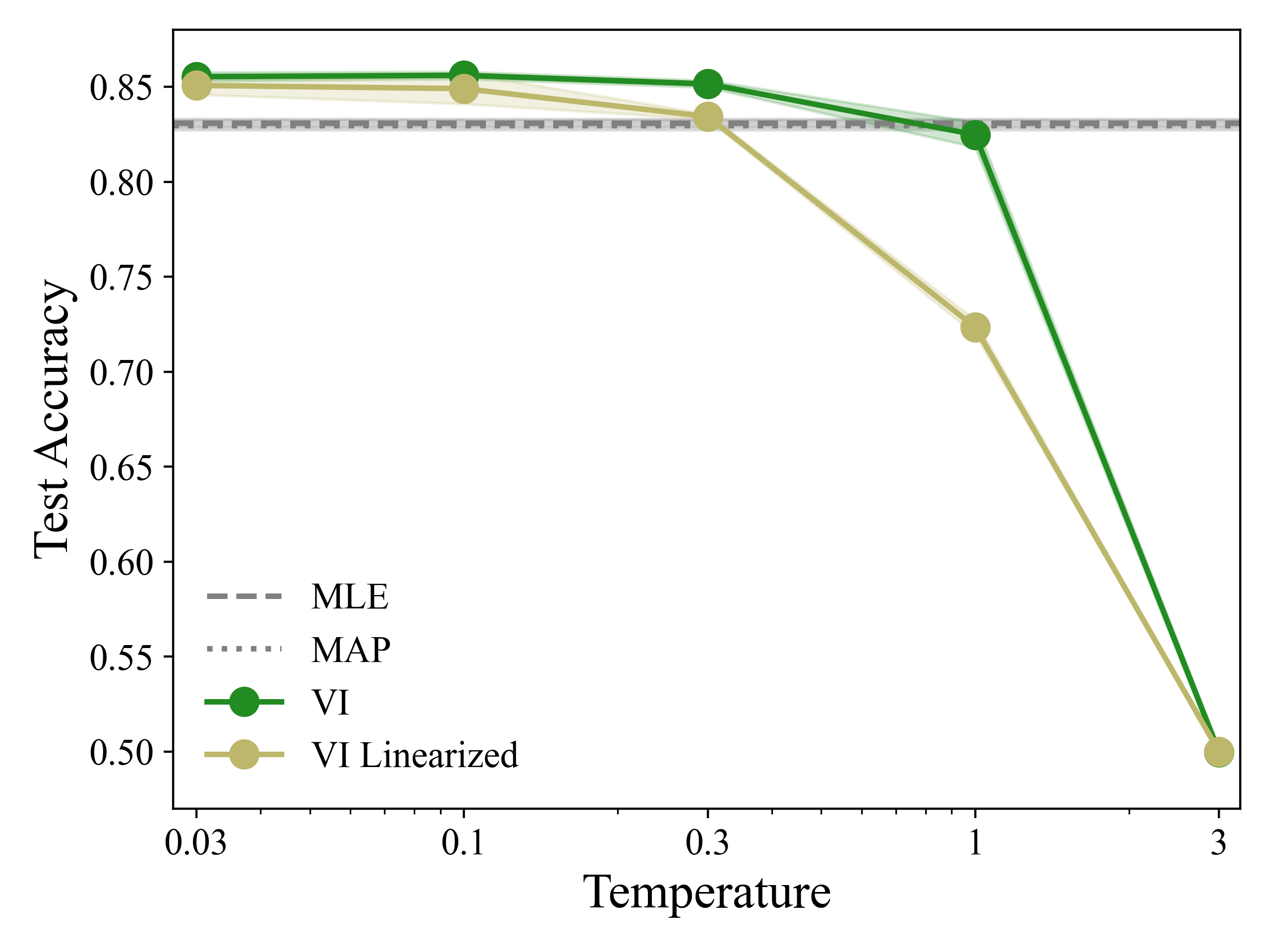}
    \end{subfigure}
    \hfill 
    \begin{subfigure}[b]{0.32\textwidth}
        \includegraphics[width=\textwidth]{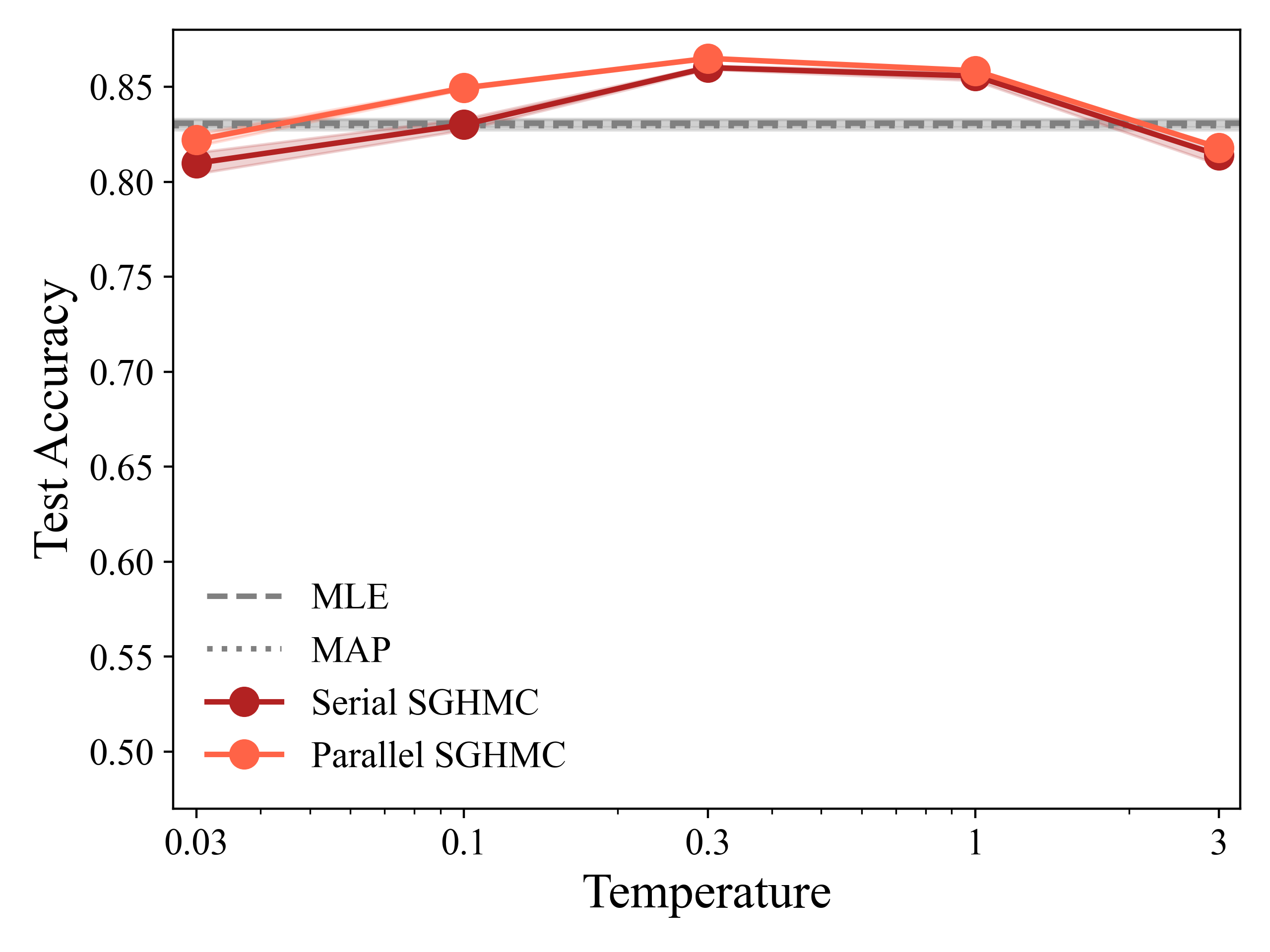}
    \end{subfigure}

    \caption{\textbf{Cold posterior test accuracies} (higher is better) for a range of \posteriors{} algorithms and temperatures for the CNN-LSTM model \citep{wenzel2020good} on the IMDB dataset \citep{maas2011learning} from Section~\ref{subsec:cpe}.}
    \label{fig:cold_posterior_acc}
\end{figure}

We here display more details on the results displayed in Figure~\ref{fig:cold_posterior}. Figure~\ref{fig:cold_posterior_with_mle} is exactly the same as Figure~\ref{fig:cold_posterior}, but with the test loss for the MLE, approach added (no prior), which severely overfits. Figure~\ref{fig:cold_posterior_acc} reports all accuracies which agree with the takeaways discussed in Section~\ref{subsec:cpe}.

We use exactly the same model CNN-LSTM from \cite{wenzel2020good, izmailov2021bayesian}, which consists of an embedding layer to convert input sequences into dense vectors, a 1D convolutional layer to capture local features, ReLU activation and max-pooling layers, an LSTM layer to capture long-term dependencies, and a fully connected layer to produce the final classification logits. The model has 2.7 million trainable parameters. Following \cite{izmailov2021bayesian} we use a diagonal Gaussian prior with all variances set to 1/40 (aside from MLE which has no prior or equivalently infinite prior variances).

We train on the IMDB \cite{maas2011learning} dataset for binary classification of positive/negative sentiment. We follow the default 50-50 split for the dataset with 25 thousand samples for training and 25 thousand samples for testing. All methods used batch size 32.

All approaches were averaged over 5 random seeds where for parallel SGHMC 5 ensembles were bootstrapped from 35 chains each with their own random seed.

We now detail the hyperparameters used for each method (except for serial SGHMC, all methods were trained for 30 epochs):
\begin{itemize}
    \item \textbf{MLE}: We train using the AdamW optimizer \citep{loshchilov2018decoupled} with all hyperparameters set to the detaults from TorchOpt \citep{Ren_TorchOpt_2023} (learning rate $10^{-3}$). 
    \item \textbf{MAP}: Same exact settings as MLE but with finite prior variances (set to 1/40 as with the other methods).
    \item \textbf{Laplace}: We take the trained MAP parameters as the mean of the Gaussian distribution. The diagonal covariances are calculated with \posteriors{}, and we compare both empirical Fisher and GGN (which is equivalent to the conditional Fisher for our model). For test set evaluation, we sample 50 samples in parameter space and $10^4$ samples in logit space for the linearized approach.
    \item \textbf{Variational Inference}: As above, we train using the AdamW optimizer \citep{loshchilov2018decoupled} with all default parameters. We apply the \textit{stick-the-landing} variance reduction \citep{roeder2017sticking} and parameterization \citep{rezende2014stochastic} tricks. We use a single sample Monte Carlo estimate at each step. We train the variational variances in log space to avoid negative variances and initialise all log standard deviations to $-3$. For test set evaluation, we sample 50 samples in parameter space and $10^4$ samples in logit space for the linearized approach.
    \item \textbf{Serial SGHMC}: We train with learning rate $0.1$ and friction $\alpha = 1$. We run for 60 epochs (which is longer as many samples are collected along a single trajectory). We apply a burn-in of 20000 thousand iterations and then collect samples every 1000 iterations, resulting in a final collection of 27 samples.
    \item \textbf{Parallel SGHMC}: Setup was exactly the same as serial except only 30 epochs were trained, and only the final parameters were taken. This was repeated across 35 different random seeds (for data shuffling and SGHMC noise) in parallel, for testing we bootstrap sampled 5 ensembles of size 15 from the 35 chains.
\end{itemize}

All cold posterior simulations were run on an NVIDIA A100, and all simulations (including repeats over 5 random seeds) take $\sim 1$ day to run.

\section{Experiment details: Continual learning with LoRA}\label{sec:details_continual_lora}

In this section, we detail details for the continual learning experiment from Section~\ref{subsec:continual_lora}.
Fully reproducible code can be found at \href{https://github.com/normal-computing/posteriors/tree/main/examples/continual_lora}{\texttt{posteriors/examples/continual\_lora}}.

Our dataset \cite{raecompressive2019} for the continual learning experiment, a collection of long books, is divided into $N$ episodes of train and test data. In the results reported in Figures~\ref{fig:continual_a} and \ref{fig:continual_b}, we use 1 book per episode, holding out the last $15\%$. We select the first twenty books from \cite{raecompressive2019} that satisfy the following criterion: 
\begin{enumerate}
    \item Published in 1900 or later
    \item Is at least 100,000 tokens, and at most 1,000,000 tokens
\end{enumerate}

We use LoRA \citep{hu2021lora} on Llama 2 \citep{touvron2023llama} to reduce the dimension of the inference task to 200k trainable parameters.  We test two baseline methods. We fine-tune the model on each episode's new data, a standard approach known to result in catastrophic forgetting, and we also implement a static offline baseline that sees all data every episode. This represents the LoRA network's total capacity but is computationally infeasible for online learning.

The prior for the next episode is the posterior from the previous episode, becoming a quadratic penalty in the loss function during gradient descent. Whereas the original work on catastrophic forgetting \citep{kirkpatrick2017overcoming} suggested using multiple penalties (incorporating data from all previous episodes into the prior), we use a single penalty following \cite{huszar2018note}. We use only the last episode's posterior, reminiscent of exact Bayesian updates. 

For each episode of data, we perform fine-tuning on the current episode's data and test on for all data episodes seen thus far. The Laplace method additionally updates the posterior based on new MAP estimates of weights, and the latest Fisher information. The posterior is used as the prior for the next episode. These steps are handled by \posteriors{}.

Additional Experiment Settings:
\begin{enumerate}
    \item We use rank $r=8$ and $\alpha = 32$ for LoRA, the standard settings \citep{hu2021lora}, and fine-tune three weight matrices in the last layer (key, query, and output projections), following the literature \citep{yang2024bayesian}.
    \item We stride over the texts so that all tokens have 2048 context tokens, given that there are 2048 previous tokens to 
    condition on.  
\end{enumerate}

Continual LoRA simulations were run on an NVIDIA A100; simulations over the 20 books take $\sim 6$ hours to run.

\section{Experiment details: Bayesian Llama 3}\label{sec:details_bayesllama}

Here we describe further details for our final experiment fine-tuning the 8B Llama 3 model \citep{llama3modelcard}. Fully reproducible code can be found at \href{https://github.com/normal-computing/posteriors/tree/main/examples/bayes_llama3}{\texttt{posteriors/examples/bayes\_llama3}}.

As previously mentioned, our experiments involve using 10 sets of ensemble models for the last attention layer of the Llama3 model, resulting in 218 million trainable parameters. We train via SGHMC sampling for 20 epochs over the TQA dataset \citep{kembhavi2017you}. For evaluation, we generated a random set of 100 scientific factual statements from ChatGPT, which we also validated by hand. Additional hyperparameter and training details are as follows: 

\begin{enumerate}
    \item We set the learning rate to be $10^{-3}$. We set alpha and beta parameters to $10^{-2}$ and 0, respectively, with the momenta initialized to 0.
    \item For the textbook content, we tokenize with a stride length of 300 and a stride overlap of size 100 while using a batch size of 10. During training, we mask the loss on the first 250 tokens and only consider the loss on the last 50 tokens. 
    \item All layers are frozen except for the final attention layer.
\end{enumerate}

\begin{figure}[ht]
    \centering
    \includegraphics[width=\textwidth]{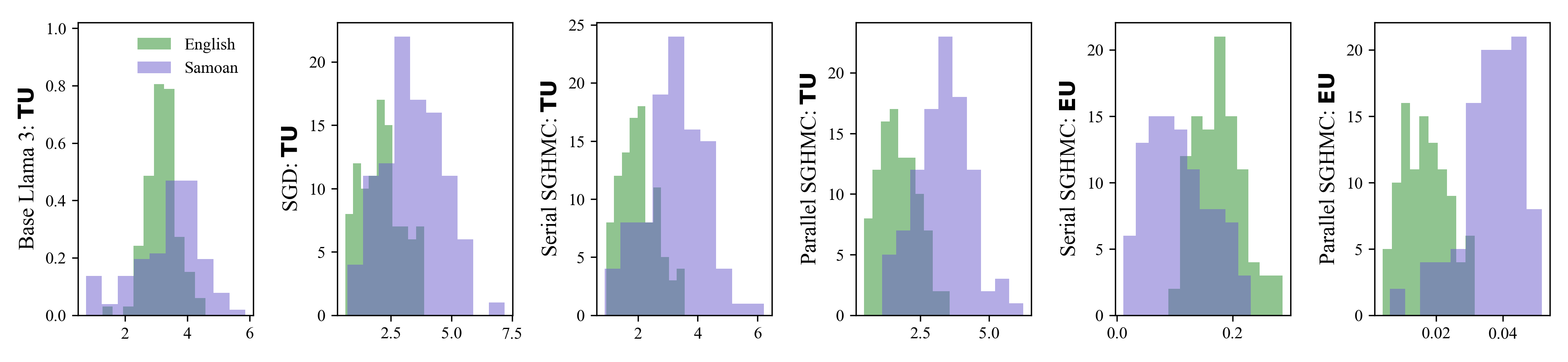}
    \caption{\textbf{Distributions of predictive uncertainties} with Llama 3 (Serial SGHMC included).}
    \label{fig:bayesllama_hists_appendix}
\end{figure}

In Table~\ref{tab:loss_with_samoan}, we extend Table~\ref{tab:auroc} to include the loss results for predicting the last word of the statements translated into Samoan. Where we can clearly see the Samoan text is significantly out of distribution. We also include one standard deviation error from 4 random seeds for SGD.

In Figure~\ref{fig:bayesllama_hists_appendix}, we extend Figure~\ref{fig:bayesllama_hists} to include results for serial SGHMC. We see that serial SGHMC fails to provide a useful measure of epistemic uncertainty, which, as discussed in Section~\ref{subsec:bayesllama}, we posit is due to serial SGHMC struggling to escape local modes and, therefore, lacking diversity in predictions - and even further, as the collected parameters are very similar (in such high dimension), this gives a misguided indication of low epistemic uncertainty (a large agreement in produced logits) that results in worse performance than using TU.

Bayesian Llama 3 simulations were run on an NVIDIA A100; training of the SGD and ensemble approaches take $\sim 16$ hours to run whilst evaluation over the 100 statements is fast.

\begin{table}
  \caption{\textbf{Out-of-distribution metrics} with Bayesian Llama 3.}\label{tab:loss_with_samoan}
  \centering
  \begin{tabular}{lcccccc}
    \toprule
    & Base Llama 3 & SGD& \multicolumn{2}{c}{Serial SGHMC}  & \multicolumn{2}{c}{Parallel SGHMC}   \\
         & \bf{TU}    &  \bf{TU}  & \: \:  \bf{TU} & \bf{EU} & \: \:  \bf{TU} & \bf{EU}  \\
    \midrule
    \textbf{AUROC} $\uparrow$ &   0.62 & 0.80 $\pm$ 0.02 & \: \: 0.85 & 0.17 & \: \: 0.90 & \bf{0.95} \\
    \textbf{Loss on English statements} $\downarrow$ &   4.68 & \textbf{4.20} $\pm$ 0.14 & \multicolumn{2}{c}{4.28} & \multicolumn{2}{c}{4.40} \\
    \textbf{Loss on Samoan statements} $\downarrow$ &   11.41 & 9.87 $\pm$ 0.05 & \multicolumn{2}{c}{9.98} & \multicolumn{2}{c}{\textbf{9.83}} \\
    \bottomrule
  \end{tabular}
\end{table}

\section{Bayesian logistic regression: Comparison and composition with Pyro}\label{sec:PI}

In this section, we use the probabilistic programming language (PPL) Pyro~\citep{bingham2019pyro} to construct a more classical Bayesian logistic regression model on the Pima Indians dataset \citep{smith1988using} which is a standard benchmark in computational statistics, see e.g. \cite{titsias2024optimal}. The model is constructed in Pyro and seamlessly converted into a \texttt{log\_posterior} function for scalable inference with \posteriors{}. These simulations demonstrate appropriate convergence of the \posteriors{} in a smaller-scale setting, compare efficiency to other frameworks and demonstrate seamless composition with a PPL in Pyro.

We compare \posteriors{}' SGHMC and VI implementations with Pyro’s automated NUTS \citep{hoffman2014no} implementation (which represents a gold standard) and a JAX \citep{jax2018github} implementation with BlackJAX \citep{cabezas2024blackjax}. All serial MCMC methods were run for 5000 samples with similar warmup and thinning configurations (noting that NUTS has an adaptive thinning procedure). VI was ran for 8000 iterations which was sufficient for convergence. For parallel SGHMC we run 5000 samples independently for 2000 iterations.

 We observe, in Figure~\ref{fig:PI_fullbatch}, that all methods (including mini-batched versions, Figure~\ref{fig:PI_minibatch}) are broadly in agreement and have roughly converged to the same posterior distribution. In Table~\ref{tab:PI}, we compare the fidelity of the posterior approximations by visualising the marginal distributions and comparing quantitatively via a kernelized Stein discrepancy \citep{gorham2015measuring}. We also display the runtimes of the approaches although in this case the model is small (8 dimensions and 768 data points) and therefore the Python overheads are relatively significant over model calls (as demonstrated by only a marginal speedup from mini-batching) and this is more so for PyTorch over JAX (although this rapidly deteriorates in the case of larger, more expensive model calls, in our experience). Nevertheless, we see that posteriors is competitive with Pyro~\citep{bingham2019pyro}. Importantly, these simulations demonstrate that \posteriors{ can easily be composed one of the most popular probabilistic programming languages in Pyro~\citep{bingham2019pyro} to perform efficient and scalable mini-batch inference in traditional Bayesian models.

\begin{figure}
    \centering
    \includegraphics[width=0.95\linewidth]{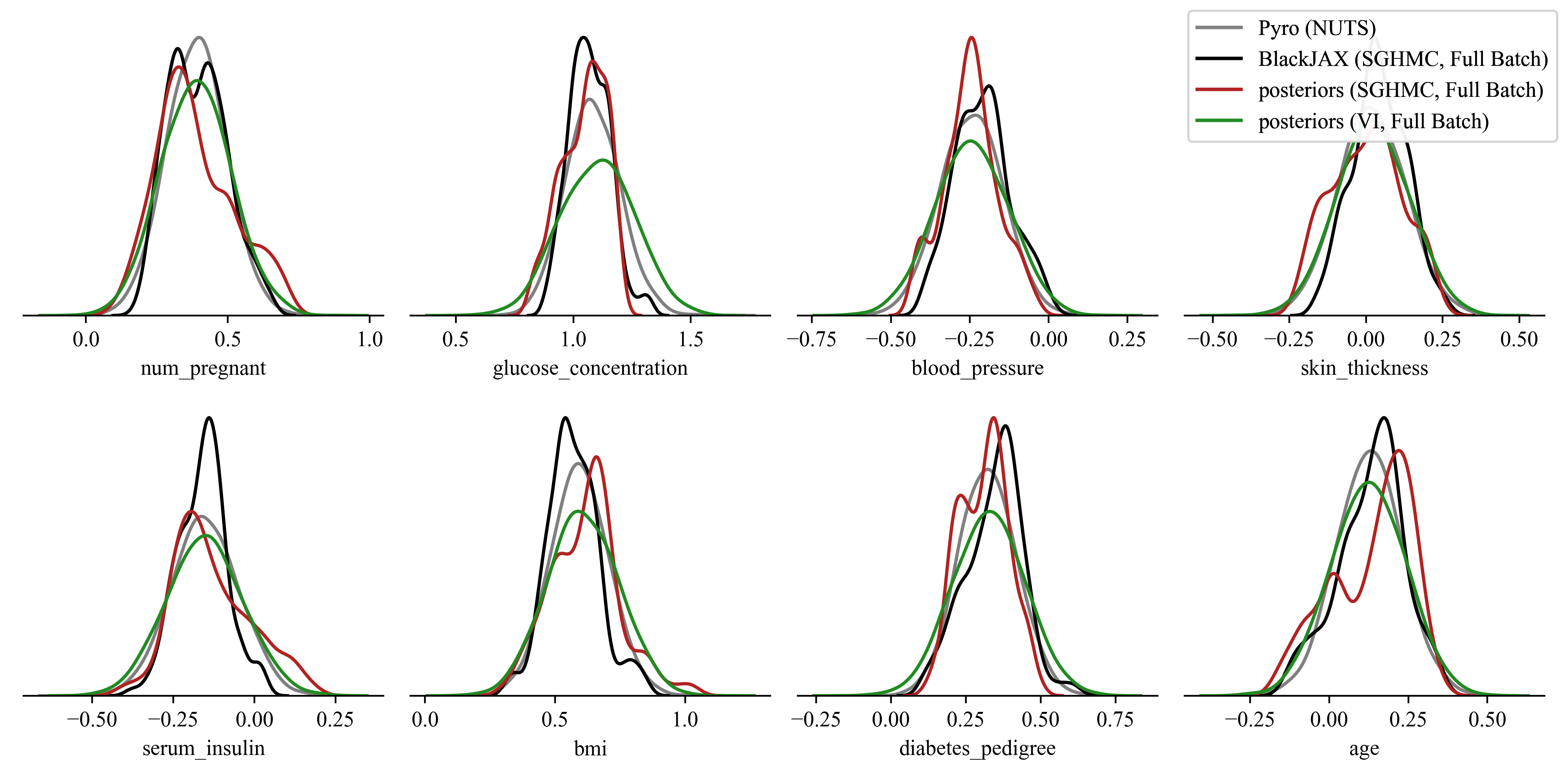}
    \caption{\textbf{Full batch marginal distributions for Bayesian logistic regression} on the Pima Indians dataset. All methods generated 5000 samples and provided approximations that are broadly in agreement.}
    \label{fig:PI_fullbatch}
\end{figure}

\begin{figure}
    \centering
    \includegraphics[width=0.95\linewidth]{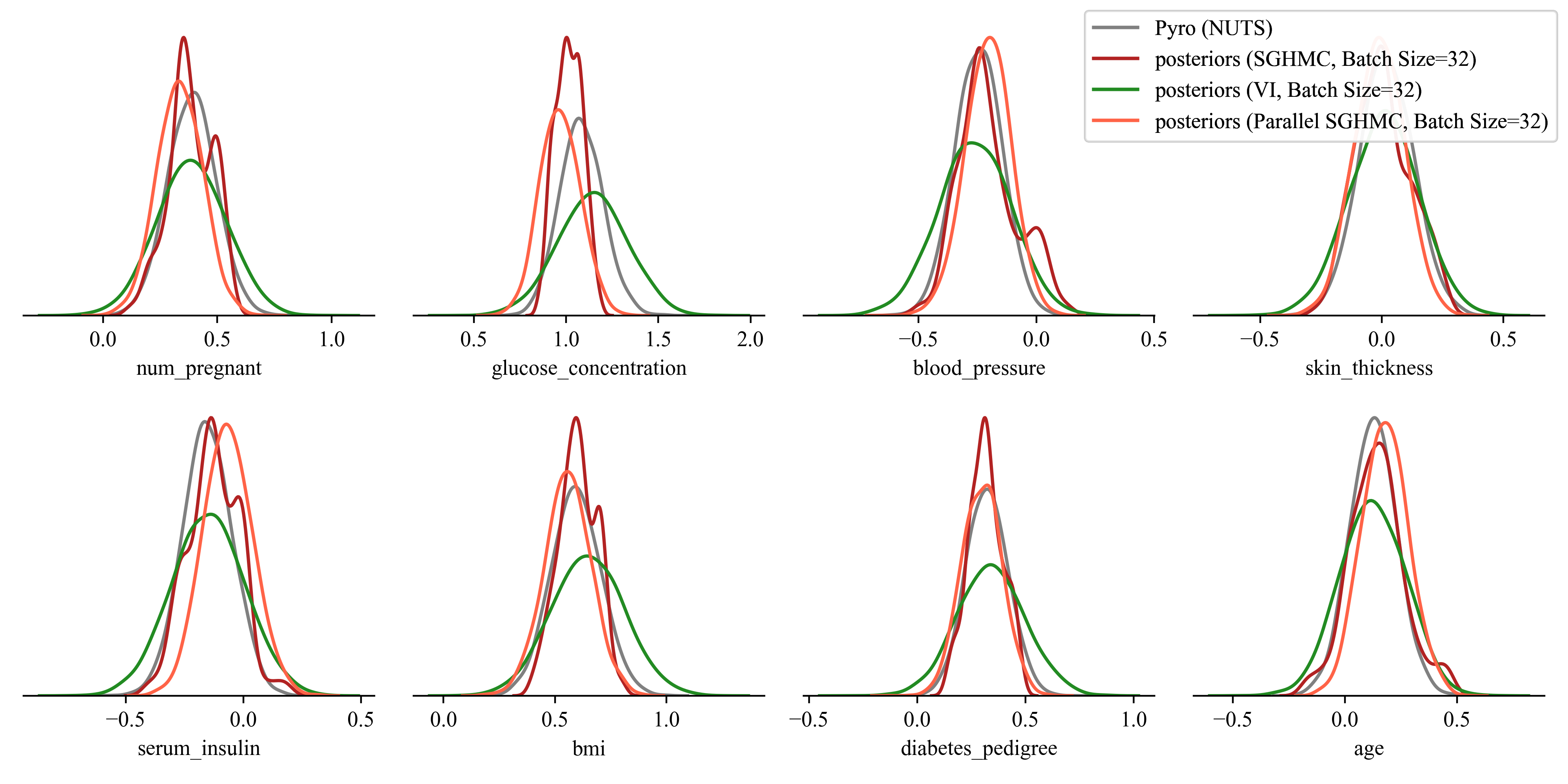}
    \caption{\textbf{Mini batch marginal distributions for Bayesian logistic regression}. Otherwise, same setup as Figure~\ref{fig:PI_fullbatch} with full batch Pyro retained as a baseline. Again all methods provide relatively consistent posterior approximations.}
    \label{fig:PI_minibatch}
\end{figure}

\begin{table}[h]
    \caption{\textbf{Metrics for Bayesian Logistic Regression.} Kernelized Stein discrepancy (KSD) \cite{gorham2015measuring} measures the distance between the samples provided by the algorithm and the true posterior via a kernel function (in this case a standard Gaussian). All results are averaged over 10 random seeds with one standard deviation displayed. {}\textsuperscript{\textdagger}The displayed parallel SGHMC time represents the time for a single chain that could be obtained with sufficient parallel resources.}
    \centering
    \begin{tabular}{lcc}
    \toprule
                                              & \textbf{KSD} $\downarrow$         & \textbf{Time} (s) $\downarrow$  \\
    \midrule
    Pyro  (NUTS \cite{hoffman2014no})          & 2.680 ± 0.001 & 15.38 ± 0.74 \\
    BlackJAX (SGHMC, Full Batch)              & 2.689 ± 0.013 & 5.29 ± 0.74  \\
    posteriors (VI, Full Batch)               & 2.623 ± 0.001 & 7.04 ± 0.23  \\
    posteriors (VI, Batch Size=32)            & 2.516 ± 0.003 & 6.52 ± 0.22  \\
    posteriors (SGHMC, Full Batch)            & 2.693 ± 0.024 & 14.37 ± 0.31 \\
    posteriors (SGHMC, Batch Size=32)         & 2.686 ± 0.012 & 13.73 ± 0.38 \\
    posteriors (Parallel SGHMC, Batch Size=32) & 2.690 ± 0.001 & 0.79 ± 0.04{}\textsuperscript{\textdagger} \\
    \bottomrule
    \label{tab:PI}
    \end{tabular}
\end{table}

\end{document}

%% file: math_commands.tex
\usepackage{amsmath,amsfonts,bm}

\def\eqref#1{equation~\ref{#1}}

\def\1{\bm{1}}

\DeclareMathAlphabet{\mathsfit}{\encodingdefault}{\sfdefault}{m}{sl}
\SetMathAlphabet{\mathsfit}{bold}{\encodingdefault}{\sfdefault}{bx}{n}

\newcommand{\E}{\mathbb{E}}

\DeclareMathOperator*{\argmax}{arg\,max}

%% file: figs/bayes_benefits_schematic.tex
\tikzset{every picture/.style={line width=0.75pt}} 

\begin{tikzpicture}[x=0.75pt,y=0.75pt,yscale=-1,xscale=1, baseline={(0,0)}]

\draw  [draw opacity=0][fill={rgb, 255:red, 154; green, 203; blue, 255 }  ,fill opacity=1 ] (27,223.5) .. controls (27,208.5) and (173,282.33) .. (203,267) .. controls (233,251.67) and (243,128.5) .. (291,114.5) .. controls (339,100.5) and (364,143.5) .. (426,167.5) .. controls (488,191.5) and (518,205.5) .. (547,189.5) .. controls (576,173.5) and (665,-12.83) .. (665,2.5) .. controls (665,17.83) and (665,304.5) .. (665,313.5) .. controls (665,322.5) and (490,232.5) .. (436,224.5) .. controls (382,216.5) and (393,227.5) .. (344,265.5) .. controls (295,303.5) and (280,290.5) .. (232,305.5) .. controls (184,320.5) and (30,411.5) .. (30,401.5) .. controls (30,391.5) and (27,238.5) .. (27,223.5) -- cycle ;
\draw [color={rgb, 255:red, 119; green, 119; blue, 119 }  ,draw opacity=1 ][line width=4.5]    (28,312) .. controls (76,306) and (184,315) .. (224,285) .. controls (279,234.5) and (286,136) .. (388,180) .. controls (614,289) and (628.33,142.17) .. (666.33,88.33) ;
\draw  [color={rgb, 255:red, 255; green, 64; blue, 64 }  ,draw opacity=1 ][line width=3]  (180,290.5) -- (197,290.5)(188.5,282) -- (188.5,299) ;
\draw  [color={rgb, 255:red, 255; green, 64; blue, 64 }  ,draw opacity=1 ][line width=3]  (216,294.5) -- (233,294.5)(224.5,286) -- (224.5,303) ;
\draw  [color={rgb, 255:red, 255; green, 64; blue, 64 }  ,draw opacity=1 ][line width=3]  (213,262.5) -- (230,262.5)(221.5,254) -- (221.5,271) ;
\draw  [color={rgb, 255:red, 255; green, 64; blue, 64 }  ,draw opacity=1 ][line width=3]  (241,280.5) -- (258,280.5)(249.5,272) -- (249.5,289) ;
\draw  [color={rgb, 255:red, 255; green, 64; blue, 64 }  ,draw opacity=1 ][line width=3]  (379,171.5) -- (396,171.5)(387.5,163) -- (387.5,180) ;
\draw  [color={rgb, 255:red, 255; green, 64; blue, 64 }  ,draw opacity=1 ][line width=3]  (396,198.5) -- (413,198.5)(404.5,190) -- (404.5,207) ;
\draw  [color={rgb, 255:red, 255; green, 64; blue, 64 }  ,draw opacity=1 ][line width=3]  (172,309.5) -- (189,309.5)(180.5,301) -- (180.5,318) ;
\draw  [color={rgb, 255:red, 255; green, 64; blue, 64 }  ,draw opacity=1 ][line width=3]  (426,185.5) -- (443,185.5)(434.5,177) -- (434.5,194) ;
\draw  [color={rgb, 255:red, 255; green, 64; blue, 64 }  ,draw opacity=1 ][line width=3]  (422,207.5) -- (439,207.5)(430.5,199) -- (430.5,216) ;
\draw  [color={rgb, 255:red, 255; green, 64; blue, 64 }  ,draw opacity=1 ][line width=3]  (440,214.5) -- (457,214.5)(448.5,206) -- (448.5,223) ;
\draw  [color={rgb, 255:red, 255; green, 64; blue, 64 }  ,draw opacity=1 ][line width=3]  (453,194.5) -- (470,194.5)(461.5,186) -- (461.5,203) ;
\draw  [color={rgb, 255:red, 255; green, 64; blue, 64 }  ,draw opacity=1 ][line width=3]  (467,221.5) -- (484,221.5)(475.5,213) -- (475.5,230) ;

\draw  [color={rgb, 255:red, 255; green, 255; blue, 255 }  ,draw opacity=1 ][fill={rgb, 255:red, 255; green, 255; blue, 255 }  ,fill opacity=1 ] (668,0) -- (663,0) -- (663,350) -- (668,350) -- cycle ;

\draw (145,339) node [anchor=north west][inner sep=0.75pt] [font=\Huge] [align=left] {\begin{minipage}[lt]{173.49pt}\setlength\topsep{0pt}
\begin{center}
\textbf{{\Huge Confident}}\\\textbf{{\Huge in-distribution}}
\end{center}

\end{minipage}};
\draw (120,11) node [anchor=north west][inner sep=0.75pt]  [font=\Huge] [align=left] {\begin{minipage}[lt]{222.85pt}\setlength\topsep{0pt}
\begin{center}
\textbf{{\Huge Uncertain}}\\\textbf{{\Huge out-of-distribution}}
\end{center}

\end{minipage}};

\end{tikzpicture}

\begin{tikzpicture}[x=0.75pt,y=0.75pt,yscale=-1,xscale=1, baseline={(0,0)}]
uncomment if require: \path (0,458); 

\draw  [draw opacity=0][fill={rgb, 255:red, 154; green, 203; blue, 255 }  ,fill opacity=1 ] (27,223.5) .. controls (27,208.5) and (173,282.33) .. (203,267) .. controls (233,251.67) and (243,128.5) .. (291,114.5) .. controls (339,100.5) and (364,143.5) .. (426,167.5) .. controls (488,191.5) and (530,202.5) .. (555.5,191.75) .. controls (581,181) and (589,175.75) .. (609,145.75) .. controls (629,115.75) and (665.5,46.58) .. (665.5,54.25) .. controls (665.5,61.92) and (663,171.75) .. (663,175.75) .. controls (663,179.75) and (644,180.25) .. (610.5,217.25) .. controls (577,254.25) and (463,228.5) .. (436,224.5) .. controls (409,220.5) and (393,227.5) .. (344,265.5) .. controls (295,303.5) and (280,290.5) .. (232,305.5) .. controls (184,320.5) and (30,411.5) .. (30,401.5) .. controls (30,391.5) and (27,238.5) .. (27,223.5) -- cycle ;
\draw [color={rgb, 255:red, 119; green, 119; blue, 119 }  ,draw opacity=1 ][line width=4.5]    (28,312) .. controls (76,306) and (184,315) .. (224,285) .. controls (279,234.5) and (286,136) .. (388,180) .. controls (614,289) and (628.33,142.17) .. (666.33,88.33) ;
\draw  [color={rgb, 255:red, 255; green, 64; blue, 64 }  ,draw opacity=1 ][line width=3]  (180,290.5) -- (197,290.5)(188.5,282) -- (188.5,299) ;
\draw  [color={rgb, 255:red, 255; green, 64; blue, 64 }  ,draw opacity=1 ][line width=3]  (216,294.5) -- (233,294.5)(224.5,286) -- (224.5,303) ;
\draw  [color={rgb, 255:red, 255; green, 64; blue, 64 }  ,draw opacity=1 ][line width=3]  (213,262.5) -- (230,262.5)(221.5,254) -- (221.5,271) ;
\draw  [color={rgb, 255:red, 255; green, 64; blue, 64 }  ,draw opacity=1 ][line width=3]  (241,280.5) -- (258,280.5)(249.5,272) -- (249.5,289) ;
\draw  [color={rgb, 255:red, 255; green, 64; blue, 64 }  ,draw opacity=1 ][line width=3]  (379,171.5) -- (396,171.5)(387.5,163) -- (387.5,180) ;
\draw  [color={rgb, 255:red, 255; green, 64; blue, 64 }  ,draw opacity=1 ][line width=3]  (396,198.5) -- (413,198.5)(404.5,190) -- (404.5,207) ;
\draw  [color={rgb, 255:red, 255; green, 64; blue, 64 }  ,draw opacity=1 ][line width=3]  (172,309.5) -- (189,309.5)(180.5,301) -- (180.5,318) ;
\draw  [color={rgb, 255:red, 255; green, 64; blue, 64 }  ,draw opacity=1 ][line width=3]  (426,185.5) -- (443,185.5)(434.5,177) -- (434.5,194) ;
\draw  [color={rgb, 255:red, 255; green, 64; blue, 64 }  ,draw opacity=1 ][line width=3]  (422,207.5) -- (439,207.5)(430.5,199) -- (430.5,216) ;
\draw  [color={rgb, 255:red, 255; green, 64; blue, 64 }  ,draw opacity=1 ][line width=3]  (440,214.5) -- (457,214.5)(448.5,206) -- (448.5,223) ;
\draw  [color={rgb, 255:red, 255; green, 64; blue, 64 }  ,draw opacity=1 ][line width=3]  (453,194.5) -- (470,194.5)(461.5,186) -- (461.5,203) ;
\draw  [color={rgb, 255:red, 255; green, 64; blue, 64 }  ,draw opacity=1 ][line width=3]  (467,220.5) -- (484,220.5)(475.5,212) -- (475.5,229) ;
\draw  [color={rgb, 255:red, 217; green, 82; blue, 209 }  ,draw opacity=1 ][line width=3]  (568.5,193.5) -- (585.5,193.5)(577,185) -- (577,202) ;
\draw  [color={rgb, 255:red, 217; green, 82; blue, 209 }  ,draw opacity=1 ][line width=3]  (592.5,208.5) -- (609.5,208.5)(601,200) -- (601,217) ;
\draw  [color={rgb, 255:red, 217; green, 82; blue, 209 }  ,draw opacity=1 ][line width=3]  (600,175) -- (617,175)(608.5,166.5) -- (608.5,183.5) ;
\draw  [color={rgb, 255:red, 217; green, 82; blue, 209 }  ,draw opacity=1 ][line width=3]  (627.5,167) -- (644.5,167)(636,158.5) -- (636,175.5) ;

\draw  [color={rgb, 255:red, 255; green, 255; blue, 255 }  ,draw opacity=1 ][fill={rgb, 255:red, 255; green, 255; blue, 255 }  ,fill opacity=1 ] (668,50.75) -- (663,50.75) -- (663,181.75) -- (668,181.75) -- cycle ;

\draw (434,36) node [anchor=north west][inner sep=0.75pt]  [font=\Huge] [align=left] {\begin{minipage}[lt]{142.52pt}\setlength\topsep{0pt}
\begin{center}
\textbf{{\Huge Learns new}}\\\textbf{{\Huge data}}
\end{center}

\end{minipage}};
\draw (184,325) node [anchor=north west][inner sep=0.75pt]  [font=\Huge] [align=left] {\begin{minipage}[lt]{167.93pt}\setlength\topsep{0pt}
\begin{center}
\textbf{{\Huge Remembers}}\\\textbf{{\Huge previous data}}
\end{center}

\end{minipage}};

\end{tikzpicture}

\begin{tikzpicture}[x=0.75pt,y=0.75pt,yscale=-1,xscale=1, baseline={(0,0)}]
uncomment if require: \path (0,458); 

\draw  [draw opacity=0][fill={rgb, 255:red, 108; green, 122; blue, 247 }  ,fill opacity=1 ] (27,223.5) .. controls (27,208.5) and (173,282.33) .. (203,267) .. controls (233,251.67) and (243,128.5) .. (291,114.5) .. controls (339,100.5) and (364,143.5) .. (426,167.5) .. controls (488,191.5) and (530,202.5) .. (555.5,191.75) .. controls (581,181) and (589,175.75) .. (609,145.75) .. controls (629,115.75) and (665.5,46.58) .. (665.5,54.25) .. controls (665.5,61.92) and (663,171.75) .. (663,175.75) .. controls (663,179.75) and (644,180.25) .. (610.5,217.25) .. controls (577,254.25) and (463,228.5) .. (436,224.5) .. controls (409,220.5) and (393,227.5) .. (344,265.5) .. controls (295,303.5) and (280,290.5) .. (232,305.5) .. controls (184,320.5) and (30,411.5) .. (30,401.5) .. controls (30,391.5) and (27,238.5) .. (27,223.5) -- cycle ;
\draw  [draw opacity=0][fill={rgb, 255:red, 185; green, 198; blue, 255 }  ,fill opacity=1 ] (28.33,287) .. controls (28.33,282.33) and (173,282.33) .. (203,267) .. controls (233,251.67) and (239.67,182.33) .. (303,149.67) .. controls (366.33,117) and (488.67,217.33) .. (549,200) .. controls (609.33,182.67) and (665,51.67) .. (665,67) .. controls (665,82.33) and (664.75,123.75) .. (664.75,128.75) .. controls (664.75,133.75) and (648,202) .. (575,228.25) .. controls (502,254.5) and (369.67,186.33) .. (333,203) .. controls (296.33,219.67) and (305,270.75) .. (223.5,308.75) .. controls (142,346.75) and (29,346.33) .. (29,340) .. controls (29,333.67) and (28.33,291.67) .. (28.33,287) -- cycle ;
\draw [color={rgb, 255:red, 119; green, 119; blue, 119 }  ,draw opacity=1 ][line width=4.5]    (28,312) .. controls (76,306) and (184,315) .. (224,285) .. controls (279,234.5) and (286,136) .. (388,180) .. controls (614,289) and (628.33,142.17) .. (666.33,88.33) ;
\draw  [color={rgb, 255:red, 255; green, 64; blue, 64 }  ,draw opacity=1 ][line width=3]  (180,289.5) -- (197,289.5)(188.5,281) -- (188.5,298) ;
\draw  [color={rgb, 255:red, 255; green, 64; blue, 64 }  ,draw opacity=1 ][line width=3]  (216,294.5) -- (233,294.5)(224.5,286) -- (224.5,303) ;
\draw  [color={rgb, 255:red, 255; green, 64; blue, 64 }  ,draw opacity=1 ][line width=3]  (213,262.5) -- (230,262.5)(221.5,254) -- (221.5,271) ;
\draw  [color={rgb, 255:red, 255; green, 64; blue, 64 }  ,draw opacity=1 ][line width=3]  (241,280.5) -- (258,280.5)(249.5,272) -- (249.5,289) ;
\draw  [color={rgb, 255:red, 255; green, 64; blue, 64 }  ,draw opacity=1 ][line width=3]  (379,171.5) -- (396,171.5)(387.5,163) -- (387.5,180) ;
\draw  [color={rgb, 255:red, 255; green, 64; blue, 64 }  ,draw opacity=1 ][line width=3]  (396,198.5) -- (413,198.5)(404.5,190) -- (404.5,207) ;
\draw  [color={rgb, 255:red, 255; green, 64; blue, 64 }  ,draw opacity=1 ][line width=3]  (172,309.5) -- (189,309.5)(180.5,301) -- (180.5,318) ;
\draw  [color={rgb, 255:red, 255; green, 64; blue, 64 }  ,draw opacity=1 ][line width=3]  (426,185.5) -- (443,185.5)(434.5,177) -- (434.5,194) ;
\draw  [color={rgb, 255:red, 255; green, 64; blue, 64 }  ,draw opacity=1 ][line width=3]  (422,207.5) -- (439,207.5)(430.5,199) -- (430.5,216) ;
\draw  [color={rgb, 255:red, 255; green, 64; blue, 64 }  ,draw opacity=1 ][line width=3]  (440,214.5) -- (457,214.5)(448.5,206) -- (448.5,223) ;
\draw  [color={rgb, 255:red, 255; green, 64; blue, 64 }  ,draw opacity=1 ][line width=3]  (453,194.5) -- (470,194.5)(461.5,186) -- (461.5,203) ;
\draw  [color={rgb, 255:red, 255; green, 64; blue, 64 }  ,draw opacity=1 ][line width=3]  (467,220.5) -- (484,220.5)(475.5,212) -- (475.5,229) ;
\draw  [color={rgb, 255:red, 255; green, 64; blue, 64 }  ,draw opacity=1 ][line width=3]  (569.5,193.5) -- (586.5,193.5)(578,185) -- (578,202) ;
\draw  [color={rgb, 255:red, 255; green, 64; blue, 64 }  ,draw opacity=1 ][line width=3]  (592.5,208.5) -- (609.5,208.5)(601,200) -- (601,217) ;
\draw  [color={rgb, 255:red, 255; green, 64; blue, 64 }  ,draw opacity=1 ][line width=3]  (600,175) -- (617,175)(608.5,166.5) -- (608.5,183.5) ;
\draw  [color={rgb, 255:red, 255; green, 64; blue, 64 }  ,draw opacity=1 ][line width=3]  (627.5,167) -- (644.5,167)(636,158.5) -- (636,175.5) ;

\draw  [color={rgb, 255:red, 255; green, 255; blue, 255 }  ,draw opacity=1 ][fill={rgb, 255:red, 255; green, 255; blue, 255 }  ,fill opacity=1 ] (668,50.75) -- (663,50.75) -- (663,181.75) -- (668,181.75) -- cycle ;

\draw (93.25,325) node  [font=\huge] [align=left] {\begin{minipage}[lt]{91.46pt}\setlength\topsep{0pt}
\textbf{Aleatoric}
\end{minipage}};
\draw (89.92,269) node  [font=\huge] [align=left] {\begin{minipage}[lt]{91.91pt}\setlength\topsep{0pt}
\textbf{\textcolor[rgb]{1,1,1}{Epistemic}}
\end{minipage}};
\draw (255,320) node [anchor=north west][inner sep=0.75pt]  [font=\Huge] [align=left] {\begin{minipage}[lt]{273.84pt}\setlength\topsep{0pt}
\begin{center}
\textbf{{\Huge Decompose predictive}}\\\textbf{{\Huge uncertainty}}
\end{center}

\end{minipage}};

\end{tikzpicture}

%% file: main_camera_ready.bbl
\begin{thebibliography}{93}
\providecommand{\natexlab}[1]{#1}
\providecommand{\url}[1]{\texttt{#1}}
\expandafter\ifx\csname urlstyle\endcsname\relax
  \providecommand{\doi}[1]{doi: #1}\else
  \providecommand{\doi}{doi: \begingroup \urlstyle{rm}\Url}\fi

\bibitem[AI@Meta(2024)]{llama3modelcard}
AI@Meta.
\newblock Llama 3 model card.
\newblock 2024.
\newblock URL \url{https://github.com/meta-llama/llama3/blob/main/MODEL_CARD.md}.

\bibitem[Aitchison(2020)]{aitchison2020statistical}
L.~Aitchison.
\newblock A statistical theory of cold posteriors in deep neural networks.
\newblock \emph{arXiv preprint arXiv:2008.05912}, 2020.

\bibitem[Alexos et~al.(2022)Alexos, Boyd, and Mandt]{alexos2022structured}
A.~Alexos, A.~J. Boyd, and S.~Mandt.
\newblock {Structured stochastic gradient MCMC}.
\newblock In \emph{International Conference on Machine Learning}, pages 414--434. PMLR, 2022.

\bibitem[Andrieu et~al.(2003)Andrieu, De~Freitas, Doucet, and Jordan]{andrieu2003introduction}
C.~Andrieu, N.~De~Freitas, A.~Doucet, and M.~I. Jordan.
\newblock {An introduction to MCMC for machine learning}.
\newblock \emph{Machine learning}, 50:\penalty0 5--43, 2003.

\bibitem[Bieringer et~al.(2023)Bieringer, Kasieczka, Steffen, and Trabs]{bieringer2023adammcmc}
S.~Bieringer, G.~Kasieczka, M.~F. Steffen, and M.~Trabs.
\newblock {AdamMCMC: Combining Metropolis Adjusted Langevin with Momentum-based Optimization}.
\newblock \emph{arXiv preprint arXiv:2312.14027}, 2023.

\bibitem[Bingham et~al.(2019)Bingham, Chen, Jankowiak, Obermeyer, Pradhan, Karaletsos, Singh, Szerlip, Horsfall, and Goodman]{bingham2019pyro}
E.~Bingham, J.~P. Chen, M.~Jankowiak, F.~Obermeyer, N.~Pradhan, T.~Karaletsos, R.~Singh, P.~A. Szerlip, P.~Horsfall, and N.~D. Goodman.
\newblock Pyro: Deep universal probabilistic programming.
\newblock \emph{J. Mach. Learn. Res.}, 20:\penalty0 28:1--28:6, 2019.
\newblock URL \url{http://jmlr.org/papers/v20/18-403.html}.

\bibitem[Blei et~al.(2017)Blei, Kucukelbir, and McAuliffe]{blei2017variational}
D.~M. Blei, A.~Kucukelbir, and J.~D. McAuliffe.
\newblock Variational inference: A review for statisticians.
\newblock \emph{Journal of the American statistical Association}, 112\penalty0 (518):\penalty0 859--877, 2017.

\bibitem[Blundell et~al.(2015)Blundell, Cornebise, Kavukcuoglu, and Wierstra]{blundell2015weight}
C.~Blundell, J.~Cornebise, K.~Kavukcuoglu, and D.~Wierstra.
\newblock Weight uncertainty in neural network.
\newblock In \emph{International conference on machine learning}, pages 1613--1622. PMLR, 2015.

\bibitem[Bradbury et~al.(2018)Bradbury, Frostig, Hawkins, Johnson, Leary, Maclaurin, Necula, Paszke, Vander{P}las, Wanderman-{M}ilne, and Zhang]{jax2018github}
J.~Bradbury, R.~Frostig, P.~Hawkins, M.~J. Johnson, C.~Leary, D.~Maclaurin, G.~Necula, A.~Paszke, J.~Vander{P}las, S.~Wanderman-{M}ilne, and Q.~Zhang.
\newblock {JAX}: composable transformations of {P}ython+{N}um{P}y programs, 2018.
\newblock URL \url{http://github.com/google/jax}.

\bibitem[Cabezas et~al.(2024)Cabezas, Corenflos, Lao, Louf, Carnec, Chaudhari, Cohn-Gordon, Coullon, Deng, Duffield, Durán-Martín, Elantkowski, Foreman-Mackey, Gregori, Iguaran, Kumar, Lysy, Murphy, Orduz, Patel, Wang, and Zinkov]{cabezas2024blackjax}
A.~Cabezas, A.~Corenflos, J.~Lao, R.~Louf, A.~Carnec, K.~Chaudhari, R.~Cohn-Gordon, J.~Coullon, W.~Deng, S.~Duffield, G.~Durán-Martín, M.~Elantkowski, D.~Foreman-Mackey, M.~Gregori, C.~Iguaran, R.~Kumar, M.~Lysy, K.~Murphy, J.~C. Orduz, K.~Patel, X.~Wang, and R.~Zinkov.
\newblock {BlackJAX: Composable Bayesian inference in JAX}, 2024.

\bibitem[Carpenter et~al.(2017)Carpenter, Gelman, Hoffman, Lee, Goodrich, Betancourt, Brubaker, Guo, Li, and Riddell]{carpenter2017stan}
B.~Carpenter, A.~Gelman, M.~D. Hoffman, D.~Lee, B.~Goodrich, M.~Betancourt, M.~Brubaker, J.~Guo, P.~Li, and A.~Riddell.
\newblock Stan: A probabilistic programming language.
\newblock \emph{Journal of statistical software}, 76\penalty0 (1), 2017.

\bibitem[Chen et~al.(2014)Chen, Fox, and Guestrin]{chen2014stochastic}
T.~Chen, E.~Fox, and C.~Guestrin.
\newblock {Stochastic gradient Hamiltonian Monte Carlo}.
\newblock In \emph{International conference on machine learning}, pages 1683--1691. PMLR, 2014.

\bibitem[Chollet et~al.(2015)]{chollet2015keras}
F.~Chollet et~al.
\newblock Keras.
\newblock \url{https://github.com/fchollet/keras}, 2015.

\bibitem[Cowles and Carlin(1996)]{cowles1996markov}
M.~K. Cowles and B.~P. Carlin.
\newblock Markov chain monte carlo convergence diagnostics: a comparative review.
\newblock \emph{Journal of the American statistical Association}, 91\penalty0 (434):\penalty0 883--904, 1996.

\bibitem[Daxberger et~al.(2021)Daxberger, Kristiadi, Immer, Eschenhagen, Bauer, and Hennig]{daxberger2021laplace}
E.~Daxberger, A.~Kristiadi, A.~Immer, R.~Eschenhagen, M.~Bauer, and P.~Hennig.
\newblock Laplace redux-effortless {B}ayesian deep learning.
\newblock \emph{Advances in Neural Information Processing Systems}, 34:\penalty0 20089--20103, 2021.

\bibitem[DeepMind et~al.(2020)DeepMind, Babuschkin, Baumli, Bell, Bhupatiraju, Bruce, Buchlovsky, Budden, Cai, Clark, Danihelka, Dedieu, Fantacci, Godwin, Jones, Hemsley, Hennigan, Hessel, Hou, Kapturowski, Keck, Kemaev, King, Kunesch, Martens, Merzic, Mikulik, Norman, Papamakarios, Quan, Ring, Ruiz, Sanchez, Sartran, Schneider, Sezener, Spencer, Srinivasan, Stanojevi\'{c}, Stokowiec, Wang, Zhou, and Viola]{deepmind2020jax}
DeepMind, I.~Babuschkin, K.~Baumli, A.~Bell, S.~Bhupatiraju, J.~Bruce, P.~Buchlovsky, D.~Budden, T.~Cai, A.~Clark, I.~Danihelka, A.~Dedieu, C.~Fantacci, J.~Godwin, C.~Jones, R.~Hemsley, T.~Hennigan, M.~Hessel, S.~Hou, S.~Kapturowski, T.~Keck, I.~Kemaev, M.~King, M.~Kunesch, L.~Martens, H.~Merzic, V.~Mikulik, T.~Norman, G.~Papamakarios, J.~Quan, R.~Ring, F.~Ruiz, A.~Sanchez, L.~Sartran, R.~Schneider, E.~Sezener, S.~Spencer, S.~Srinivasan, M.~Stanojevi\'{c}, W.~Stokowiec, L.~Wang, G.~Zhou, and F.~Viola.
\newblock The {D}eep{M}ind {JAX} {E}cosystem, 2020.
\newblock URL \url{http://github.com/google-deepmind}.

\bibitem[Detommaso et~al.(2023)Detommaso, Gasparin, Donini, Seeger, Wilson, and Archambeau]{detommaso2023fortuna}
G.~Detommaso, A.~Gasparin, M.~Donini, M.~Seeger, A.~G. Wilson, and C.~Archambeau.
\newblock Fortuna: A library for uncertainty quantification in deep learning.
\newblock \emph{arXiv preprint arXiv:2302.04019}, 2023.

\bibitem[Ding et~al.(2014)Ding, Fang, Babbush, Chen, Skeel, and Neven]{ding2014bayesian}
N.~Ding, Y.~Fang, R.~Babbush, C.~Chen, R.~D. Skeel, and H.~Neven.
\newblock Bayesian sampling using stochastic gradient thermostats.
\newblock \emph{Advances in neural information processing systems}, 27, 2014.

\bibitem[Duffield(2021)]{mocat2021}
S.~Duffield.
\newblock {mocat: All things Monte Carlo, written in JAX.}, 2021.
\newblock URL \url{http://github.com/SamDuffield/mocat}.

\bibitem[Duffield and Singh(2022)]{duffield2022quasi}
S.~Duffield and S.~S. Singh.
\newblock {Quasi-Newton Sequential Monte Carlo}.
\newblock \emph{arXiv preprint arXiv:2211.12580}, 2022.

\bibitem[Duffield et~al.(2024)Duffield, Power, and Rimella]{duffield2024state}
S.~Duffield, S.~Power, and L.~Rimella.
\newblock A state-space perspective on modelling and inference for online skill rating.
\newblock \emph{Journal of the Royal Statistical Society Series C: Applied Statistics}, 73\penalty0 (5):\penalty0 1262--1282, 2024.

\bibitem[Duran-Martin et~al.(2022)Duran-Martin, Kara, and Murphy]{duran2022efficient}
G.~Duran-Martin, A.~Kara, and K.~Murphy.
\newblock {Efficient online Bayesian inference for neural bandits}.
\newblock In \emph{International Conference on Artificial Intelligence and Statistics}, pages 6002--6021. PMLR, 2022.

\bibitem[Eschenhagen et~al.(2024)Eschenhagen, Immer, Turner, Schneider, and Hennig]{eschenhagen2024kronecker}
R.~Eschenhagen, A.~Immer, R.~Turner, F.~Schneider, and P.~Hennig.
\newblock Kronecker-factored approximate curvature for modern neural network architectures.
\newblock \emph{Advances in Neural Information Processing Systems}, 36, 2024.

\bibitem[Falcon and {The PyTorch Lightning team}(2019)]{Falcon_PyTorch_Lightning_2019}
W.~Falcon and {The PyTorch Lightning team}.
\newblock {PyTorch Lightning}, Mar. 2019.
\newblock URL \url{https://github.com/Lightning-AI/lightning}.

\bibitem[Foong et~al.(2019)Foong, Li, Hern{\'a}ndez-Lobato, and Turner]{foong2019between}
A.~Y. Foong, Y.~Li, J.~M. Hern{\'a}ndez-Lobato, and R.~E. Turner.
\newblock {`In-Between' Uncertainty in Bayesian Neural Networks}.
\newblock \emph{arXiv preprint arXiv:1906.11537}, 2019.

\bibitem[Fortuin(2022)]{fortuin2022priors}
V.~Fortuin.
\newblock {Priors in Bayesian deep learning: A review}.
\newblock \emph{International Statistical Review}, 90\penalty0 (3):\penalty0 563--591, 2022.

\bibitem[Gal and Ghahramani(2016)]{gal2016dropout}
Y.~Gal and Z.~Ghahramani.
\newblock {Dropout as a Bayesian approximation: Representing model uncertainty in deep learning}.
\newblock In \emph{international conference on machine learning}, pages 1050--1059. PMLR, 2016.

\bibitem[Gal et~al.(2017)Gal, Hron, and Kendall]{gal2017concrete}
Y.~Gal, J.~Hron, and A.~Kendall.
\newblock Concrete dropout.
\newblock \emph{Advances in neural information processing systems}, 30, 2017.

\bibitem[Gelman et~al.(2020)Gelman, Vehtari, Simpson, Margossian, Carpenter, Yao, Kennedy, Gabry, B{\"u}rkner, and Modr{\'a}k]{gelman2020bayesian}
A.~Gelman, A.~Vehtari, D.~Simpson, C.~C. Margossian, B.~Carpenter, Y.~Yao, L.~Kennedy, J.~Gabry, P.-C. B{\"u}rkner, and M.~Modr{\'a}k.
\newblock Bayesian workflow.
\newblock \emph{arXiv preprint arXiv:2011.01808}, 2020.

\bibitem[Goodfellow et~al.(2013)Goodfellow, Mirza, Xiao, Courville, and Bengio]{goodfellow2013empirical}
I.~J. Goodfellow, M.~Mirza, D.~Xiao, A.~Courville, and Y.~Bengio.
\newblock An empirical investigation of catastrophic forgetting in gradient-based neural networks.
\newblock \emph{arXiv preprint arXiv:1312.6211}, 2013.

\bibitem[Gorham and Mackey(2015)]{gorham2015measuring}
J.~Gorham and L.~Mackey.
\newblock {Measuring sample quality with Stein's method}.
\newblock \emph{Advances in neural information processing systems}, 28, 2015.

\bibitem[Grosse and Martens(2016)]{grosse2016kronecker}
R.~Grosse and J.~Martens.
\newblock A kronecker-factored approximate fisher matrix for convolution layers.
\newblock In \emph{International Conference on Machine Learning}, pages 573--582. PMLR, 2016.

\bibitem[Harrison et~al.(2024)Harrison, Willes, and Snoek]{harrison2024variational}
J.~Harrison, J.~Willes, and J.~Snoek.
\newblock {Variational Bayesian last layers}.
\newblock \emph{arXiv preprint arXiv:2404.11599}, 2024.

\bibitem[Hartigan and Hartigan(1983)]{hartigan1983asymptotic}
J.~Hartigan and J.~Hartigan.
\newblock Asymptotic normality of posterior distributions.
\newblock \emph{Bayes theory}, pages 107--118, 1983.

\bibitem[Hoffman et~al.(2014)Hoffman, Gelman, et~al.]{hoffman2014no}
M.~D. Hoffman, A.~Gelman, et~al.
\newblock {The No-U-Turn sampler: adaptively setting path lengths in Hamiltonian Monte Carlo.}
\newblock \emph{J. Mach. Learn. Res.}, 15\penalty0 (1):\penalty0 1593--1623, 2014.

\bibitem[Hofman et~al.(2024)Hofman, Sale, and H{\"u}llermeier]{hofman2024quantifying}
P.~Hofman, Y.~Sale, and E.~H{\"u}llermeier.
\newblock Quantifying aleatoric and epistemic uncertainty with proper scoring rules.
\newblock \emph{arXiv preprint arXiv:2404.12215}, 2024.

\bibitem[Hu et~al.(2021)Hu, Shen, Wallis, Allen-Zhu, Li, Wang, Wang, and Chen]{hu2021lora}
E.~J. Hu, Y.~Shen, P.~Wallis, Z.~Allen-Zhu, Y.~Li, S.~Wang, L.~Wang, and W.~Chen.
\newblock {LoRA: Low-Rank Adaptation of Large Language Models}, 2021.

\bibitem[Husz{\'a}r(2018)]{huszar2018note}
F.~Husz{\'a}r.
\newblock Note on the quadratic penalties in elastic weight consolidation.
\newblock \emph{Proceedings of the National Academy of Sciences}, 115\penalty0 (11):\penalty0 E2496--E2497, 2018.

\bibitem[Immer et~al.(2021)Immer, Korzepa, and Bauer]{immer2021improving}
A.~Immer, M.~Korzepa, and M.~Bauer.
\newblock {Improving predictions of Bayesian neural nets via local linearization}.
\newblock In \emph{International conference on artificial intelligence and statistics}, pages 703--711. PMLR, 2021.

\bibitem[Izmailov et~al.(2018)Izmailov, Podoprikhin, Garipov, Vetrov, and Wilson]{izmailov2018averaging}
P.~Izmailov, D.~Podoprikhin, T.~Garipov, D.~Vetrov, and A.~G. Wilson.
\newblock Averaging weights leads to wider optima and better generalization.
\newblock \emph{arXiv preprint arXiv:1803.05407}, 2018.

\bibitem[Izmailov et~al.(2021)Izmailov, Vikram, Hoffman, and Wilson]{izmailov2021bayesian}
P.~Izmailov, S.~Vikram, M.~D. Hoffman, and A.~G.~G. Wilson.
\newblock {What are Bayesian neural network posteriors really like?}
\newblock In \emph{International conference on machine learning}, pages 4629--4640. PMLR, 2021.

\bibitem[Jiang et~al.(2023)Jiang, Sablayrolles, Mensch, Bamford, Chaplot, Casas, Bressand, Lengyel, Lample, Saulnier, et~al.]{jiang2023mistral}
A.~Q. Jiang, A.~Sablayrolles, A.~Mensch, C.~Bamford, D.~S. Chaplot, D.~d.~l. Casas, F.~Bressand, G.~Lengyel, G.~Lample, L.~Saulnier, et~al.
\newblock Mistral 7b.
\newblock \emph{arXiv preprint arXiv:2310.06825}, 2023.

\bibitem[Kembhavi et~al.(2017)Kembhavi, Seo, Schwenk, Choi, Farhadi, and Hajishirzi]{kembhavi2017you}
A.~Kembhavi, M.~Seo, D.~Schwenk, J.~Choi, A.~Farhadi, and H.~Hajishirzi.
\newblock Are you smarter than a sixth grader? textbook question answering for multimodal machine comprehension.
\newblock In \emph{Proceedings of the IEEE Conference on Computer Vision and Pattern recognition}, pages 4999--5007, 2017.

\bibitem[Kendall and Gal(2017)]{kendall2017uncertainties}
A.~Kendall and Y.~Gal.
\newblock {What uncertainties do we need in Bayesian deep learning for computer vision?}
\newblock \emph{Advances in neural information processing systems}, 30, 2017.

\bibitem[Kingma and Ba(2014)]{kingma2014adam}
D.~P. Kingma and J.~Ba.
\newblock Adam: A method for stochastic optimization.
\newblock \emph{arXiv preprint arXiv:1412.6980}, 2014.

\bibitem[Kirkpatrick et~al.(2017)Kirkpatrick, Pascanu, Rabinowitz, Veness, Desjardins, Rusu, Milan, Quan, Ramalho, Grabska-Barwinska, et~al.]{kirkpatrick2017overcoming}
J.~Kirkpatrick, R.~Pascanu, N.~Rabinowitz, J.~Veness, G.~Desjardins, A.~A. Rusu, K.~Milan, J.~Quan, T.~Ramalho, A.~Grabska-Barwinska, et~al.
\newblock Overcoming catastrophic forgetting in neural networks.
\newblock \emph{Proceedings of the national academy of sciences}, 114\penalty0 (13):\penalty0 3521--3526, 2017.

\bibitem[Kuhn et~al.(2023)Kuhn, Gal, and Farquhar]{kuhn2023semantic}
L.~Kuhn, Y.~Gal, and S.~Farquhar.
\newblock Semantic uncertainty: Linguistic invariances for uncertainty estimation in natural language generation.
\newblock \emph{arXiv preprint arXiv:2302.09664}, 2023.

\bibitem[Kunstner et~al.(2019)Kunstner, Hennig, and Balles]{kunstner2019limitations}
F.~Kunstner, P.~Hennig, and L.~Balles.
\newblock {Limitations of the empirical Fisher approximation for natural gradient descent}.
\newblock \emph{Advances in neural information processing systems}, 32, 2019.

\bibitem[Lakshminarayanan et~al.(2017)Lakshminarayanan, Pritzel, and Blundell]{lakshminarayanan2017simple}
B.~Lakshminarayanan, A.~Pritzel, and C.~Blundell.
\newblock Simple and scalable predictive uncertainty estimation using deep ensembles.
\newblock \emph{Advances in neural information processing systems}, 30, 2017.

\bibitem[Leimkuhler et~al.(2018)Leimkuhler, Matthews, and Weare]{leimkuhler2018ensemble}
B.~Leimkuhler, C.~Matthews, and J.~Weare.
\newblock {Ensemble preconditioning for Markov chain Monte Carlo simulation}.
\newblock \emph{Statistics and Computing}, 28:\penalty0 277--290, 2018.

\bibitem[Li et~al.(2016)Li, Chen, Carlson, and Carin]{li2016preconditioned}
C.~Li, C.~Chen, D.~Carlson, and L.~Carin.
\newblock {Preconditioned stochastic gradient Langevin dynamics for deep neural networks}.
\newblock In \emph{Proceedings of the AAAI conference on artificial intelligence}, volume~30, 2016.

\bibitem[Liu et~al.(2020)Liu, Lin, Padhy, Tran, Bedrax~Weiss, and Lakshminarayanan]{liu2020simple}
J.~Liu, Z.~Lin, S.~Padhy, D.~Tran, T.~Bedrax~Weiss, and B.~Lakshminarayanan.
\newblock Simple and principled uncertainty estimation with deterministic deep learning via distance awareness.
\newblock \emph{Advances in neural information processing systems}, 33:\penalty0 7498--7512, 2020.

\bibitem[Liu and Wang(2016)]{liu2016stein}
Q.~Liu and D.~Wang.
\newblock Stein variational gradient descent: A general purpose bayesian inference algorithm.
\newblock \emph{Advances in neural information processing systems}, 29, 2016.

\bibitem[Loshchilov and Hutter(2018)]{loshchilov2018decoupled}
I.~Loshchilov and F.~Hutter.
\newblock Decoupled weight decay regularization.
\newblock In \emph{International Conference on Learning Representations}, 2018.

\bibitem[Ma et~al.(2015)Ma, Chen, and Fox]{ma2015complete}
Y.-A. Ma, T.~Chen, and E.~Fox.
\newblock {A complete recipe for stochastic gradient MCMC}.
\newblock \emph{Advances in neural information processing systems}, 28, 2015.

\bibitem[Maas et~al.(2011)Maas, Daly, Pham, Huang, Ng, and Potts]{maas2011learning}
A.~Maas, R.~E. Daly, P.~T. Pham, D.~Huang, A.~Y. Ng, and C.~Potts.
\newblock Learning word vectors for sentiment analysis.
\newblock In \emph{Proceedings of the 49th annual meeting of the association for computational linguistics: Human language technologies}, pages 142--150, 2011.

\bibitem[MacKay(1992)]{mackay1992practical}
D.~J. MacKay.
\newblock A practical bayesian framework for backpropagation networks.
\newblock \emph{Neural computation}, 4\penalty0 (3):\penalty0 448--472, 1992.

\bibitem[Maddox et~al.(2019)Maddox, Izmailov, Garipov, Vetrov, and Wilson]{maddox2019simple}
W.~J. Maddox, P.~Izmailov, T.~Garipov, D.~P. Vetrov, and A.~G. Wilson.
\newblock {A simple baseline for Bayesian uncertainty in deep learning}.
\newblock \emph{Advances in neural information processing systems}, 32, 2019.

\bibitem[Mangrulkar et~al.(2022)Mangrulkar, Gugger, Debut, Belkada, Paul, and Bossan]{peft}
S.~Mangrulkar, S.~Gugger, L.~Debut, Y.~Belkada, S.~Paul, and B.~Bossan.
\newblock {PEFT: State-of-the-art Parameter-Efficient Fine-Tuning methods}.
\newblock \url{https://github.com/huggingface/peft}, 2022.

\bibitem[Margossian et~al.(2021)Margossian, Hoffman, Sountsov, Riou-Durand, Vehtari, and Gelman]{margossian2021nested}
C.~C. Margossian, M.~D. Hoffman, P.~Sountsov, L.~Riou-Durand, A.~Vehtari, and A.~Gelman.
\newblock {Nested $\hat{R}$: Assessing the convergence of Markov chain Monte Carlo when running many short chains}.
\newblock \emph{arXiv preprint arXiv:2110.13017}, 2021.

\bibitem[Martens(2020)]{martens2020new}
J.~Martens.
\newblock New insights and perspectives on the natural gradient method.
\newblock \emph{Journal of Machine Learning Research}, 21\penalty0 (146):\penalty0 1--76, 2020.

\bibitem[Martens and Grosse(2015)]{martens2015optimizing}
J.~Martens and R.~Grosse.
\newblock {Optimizing neural networks with Kronecker-factored approximate curvature}.
\newblock In \emph{International conference on machine learning}, pages 2408--2417. PMLR, 2015.

\bibitem[Martens et~al.(2018)Martens, Ba, and Johnson]{martens2018kronecker}
J.~Martens, J.~Ba, and M.~Johnson.
\newblock Kronecker-factored curvature approximations for recurrent neural networks.
\newblock In \emph{International Conference on Learning Representations}, 2018.

\bibitem[Martens et~al.(2010)]{martens2010deep}
J.~Martens et~al.
\newblock {Deep learning via Hessian-free optimization.}
\newblock In \emph{Icml}, volume~27, pages 735--742, 2010.

\bibitem[Neal(2012)]{neal2012bayesian}
R.~M. Neal.
\newblock \emph{Bayesian learning for neural networks}, volume 118.
\newblock Springer Science \& Business Media, 2012.

\bibitem[Nemeth and Fearnhead(2021)]{nemeth2021stochastic}
C.~Nemeth and P.~Fearnhead.
\newblock {Stochastic gradient Markov chain Monte Carlo}.
\newblock \emph{Journal of the American Statistical Association}, 116\penalty0 (533):\penalty0 433--450, 2021.

\bibitem[Nguyen et~al.(2017)Nguyen, Li, Bui, and Turner]{nguyen2017variational}
C.~V. Nguyen, Y.~Li, T.~D. Bui, and R.~E. Turner.
\newblock Variational continual learning.
\newblock \emph{arXiv preprint arXiv:1710.10628}, 2017.

\bibitem[Ollivier(2018)]{ollivier2018online}
Y.~Ollivier.
\newblock Online natural gradient as a kalman filter.
\newblock 2018.

\bibitem[Osband et~al.(2024)Osband, Wen, Asghari, Dwaracherla, Ibrahimi, Lu, and Van~Roy]{osband2024epistemic}
I.~Osband, Z.~Wen, S.~M. Asghari, V.~Dwaracherla, M.~Ibrahimi, X.~Lu, and B.~Van~Roy.
\newblock Epistemic neural networks.
\newblock \emph{Advances in Neural Information Processing Systems}, 36, 2024.

\bibitem[Paszke et~al.(2019)Paszke, Gross, Massa, Lerer, Bradbury, Chanan, Killeen, Lin, Gimelshein, Antiga, et~al.]{paszke2019pytorch}
A.~Paszke, S.~Gross, F.~Massa, A.~Lerer, J.~Bradbury, G.~Chanan, T.~Killeen, Z.~Lin, N.~Gimelshein, L.~Antiga, et~al.
\newblock Pytorch: An imperative style, high-performance deep learning library.
\newblock \emph{Advances in neural information processing systems}, 32, 2019.

\bibitem[Pavliotis(2014)]{pavliotis2014stochastic}
G.~A. Pavliotis.
\newblock Stochastic processes and applications.
\newblock \emph{Texts in Applied Mathematics}, 60, 2014.

\bibitem[Phan et~al.(2019)Phan, Pradhan, and Jankowiak]{phan2019composable}
D.~Phan, N.~Pradhan, and M.~Jankowiak.
\newblock {Composable Effects for Flexible and Accelerated Probabilistic Programming in NumPyro}.
\newblock \emph{arXiv preprint arXiv:1912.11554}, 2019.

\bibitem[Rae et~al.(2019)Rae, Potapenko, Jayakumar, Hillier, and Lillicrap]{raecompressive2019}
J.~W. Rae, A.~Potapenko, S.~M. Jayakumar, C.~Hillier, and T.~P. Lillicrap.
\newblock Compressive transformers for long-range sequence modelling.
\newblock \emph{arXiv preprint}, 2019.
\newblock URL \url{https://arxiv.org/abs/1911.05507}.

\bibitem[Ren et~al.(2023)Ren, Feng, Liu, Pan, Fu, Mai, and Yang]{Ren_TorchOpt_2023}
J.~Ren, X.~Feng, B.~Liu, X.~Pan, Y.~Fu, L.~Mai, and Y.~Yang.
\newblock {TorchOpt}, Nov. 2023.
\newblock URL \url{https://github.com/metaopt/torchopt}.

\bibitem[Rezende et~al.(2014)Rezende, Mohamed, and Wierstra]{rezende2014stochastic}
D.~J. Rezende, S.~Mohamed, and D.~Wierstra.
\newblock Stochastic backpropagation and approximate inference in deep generative models.
\newblock In \emph{International conference on machine learning}, pages 1278--1286. PMLR, 2014.

\bibitem[Ritter et~al.(2018)Ritter, Botev, and Barber]{ritter2018scalable}
H.~Ritter, A.~Botev, and D.~Barber.
\newblock A scalable {Laplace} approximation for neural networks.
\newblock In \emph{6th international conference on learning representations, ICLR 2018-conference track proceedings}, volume~6. International Conference on Representation Learning, 2018.

\bibitem[Robbins and Monro(1951)]{robbins1951stochastic}
H.~Robbins and S.~Monro.
\newblock A stochastic approximation method.
\newblock \emph{The annals of mathematical statistics}, pages 400--407, 1951.

\bibitem[Roeder et~al.(2017)Roeder, Wu, and Duvenaud]{roeder2017sticking}
G.~Roeder, Y.~Wu, and D.~K. Duvenaud.
\newblock Sticking the landing: Simple, lower-variance gradient estimators for variational inference.
\newblock \emph{Advances in Neural Information Processing Systems}, 30, 2017.

\bibitem[Sale et~al.(2023)Sale, Bengs, Caprio, and H{\"u}llermeier]{sale2023second}
Y.~Sale, V.~Bengs, M.~Caprio, and E.~H{\"u}llermeier.
\newblock Second-order uncertainty quantification: A distance-based approach.
\newblock \emph{arXiv preprint arXiv:2312.00995}, 2023.

\bibitem[Smith et~al.(1988)Smith, Everhart, Dickson, Knowler, and Johannes]{smith1988using}
J.~W. Smith, J.~E. Everhart, W.~Dickson, W.~C. Knowler, and R.~S. Johannes.
\newblock Using the adap learning algorithm to forecast the onset of diabetes mellitus.
\newblock In \emph{Proceedings of the annual symposium on computer application in medical care}, page 261. American Medical Informatics Association, 1988.

\bibitem[Sutskever et~al.(2013)Sutskever, Martens, Dahl, and Hinton]{sutskever2013importance}
I.~Sutskever, J.~Martens, G.~Dahl, and G.~Hinton.
\newblock On the importance of initialization and momentum in deep learning.
\newblock In \emph{International conference on machine learning}, pages 1139--1147. PMLR, 2013.

\bibitem[Titsias(2024)]{titsias2024optimal}
M.~Titsias.
\newblock {Optimal Preconditioning and Fisher Adaptive Langevin Sampling}.
\newblock \emph{Advances in Neural Information Processing Systems}, 36, 2024.

\bibitem[Touvron et~al.(2023)Touvron, Martin, Stone, Albert, Almahairi, Babaei, Bashlykov, Batra, Bhargava, Bhosale, et~al.]{touvron2023llama}
H.~Touvron, L.~Martin, K.~Stone, P.~Albert, A.~Almahairi, Y.~Babaei, N.~Bashlykov, S.~Batra, P.~Bhargava, S.~Bhosale, et~al.
\newblock Llama 2: Open foundation and fine-tuned chat models.
\newblock \emph{arXiv preprint arXiv:2307.09288}, 2023.

\bibitem[Wang and Yeung(2020)]{wang2020survey}
H.~Wang and D.-Y. Yeung.
\newblock {A survey on Bayesian deep learning}.
\newblock \emph{ACM computing surveys (csur)}, 53\penalty0 (5):\penalty0 1--37, 2020.

\bibitem[Welling and Teh(2011)]{welling2011bayesian}
M.~Welling and Y.~W. Teh.
\newblock {Bayesian learning via stochastic gradient Langevin dynamics}.
\newblock In \emph{Proceedings of the 28th international conference on machine learning (ICML-11)}, pages 681--688. Citeseer, 2011.

\bibitem[Wenzel et~al.(2020)Wenzel, Roth, Veeling, Swiatkowski, Tran, Mandt, Snoek, Salimans, Jenatton, and Nowozin]{wenzel2020good}
F.~Wenzel, K.~Roth, B.~Veeling, J.~Swiatkowski, L.~Tran, S.~Mandt, J.~Snoek, T.~Salimans, R.~Jenatton, and S.~Nowozin.
\newblock {How Good is the Bayes Posterior in Deep Neural Networks Really?}
\newblock In \emph{International Conference on Machine Learning}, pages 10248--10259. PMLR, 2020.

\bibitem[Wilson and Izmailov(2020)]{wilson2020bayesian}
A.~G. Wilson and P.~Izmailov.
\newblock Bayesian deep learning and a probabilistic perspective of generalization.
\newblock \emph{Advances in neural information processing systems}, 33:\penalty0 4697--4708, 2020.

\bibitem[Wimmer et~al.(2023)Wimmer, Sale, Hofman, Bischl, and H{\"u}llermeier]{wimmer2023quantifying}
L.~Wimmer, Y.~Sale, P.~Hofman, B.~Bischl, and E.~H{\"u}llermeier.
\newblock Quantifying aleatoric and epistemic uncertainty in machine learning: Are conditional entropy and mutual information appropriate measures?
\newblock In \emph{Uncertainty in Artificial Intelligence}, pages 2282--2292. PMLR, 2023.

\bibitem[Wolf et~al.(2020)Wolf, Debut, Sanh, Chaumond, Delangue, Moi, Cistac, Ma, Jernite, Plu, Xu, Le~Scao, Gugger, Drame, Lhoest, and Rush]{Wolf_Transformers_State-of-the-Art_Natural_2020}
T.~Wolf, L.~Debut, V.~Sanh, J.~Chaumond, C.~Delangue, A.~Moi, P.~Cistac, C.~Ma, Y.~Jernite, J.~Plu, C.~Xu, T.~Le~Scao, S.~Gugger, M.~Drame, Q.~Lhoest, and A.~M. Rush.
\newblock {Transformers: State-of-the-Art Natural Language Processing}.
\newblock pages 38--45. Association for Computational Linguistics, Oct. 2020.
\newblock URL \url{https://www.aclweb.org/anthology/2020.emnlp-demos.6}.

\bibitem[Wu et~al.(2018)Wu, Nowozin, Meeds, Turner, Hernandez-Lobato, and Gaunt]{wu2018deterministic}
A.~Wu, S.~Nowozin, E.~Meeds, R.~E. Turner, J.~M. Hernandez-Lobato, and A.~L. Gaunt.
\newblock {Deterministic variational inference for robust Bayesian neural networks}.
\newblock \emph{arXiv preprint arXiv:1810.03958}, 2018.

\bibitem[Yang et~al.(2024)Yang, Robeyns, Wang, and Aitchison]{yang2024bayesian}
A.~X. Yang, M.~Robeyns, X.~Wang, and L.~Aitchison.
\newblock Bayesian low-rank adaptation for large language models, 2024.

\bibitem[Zhang et~al.(2018)Zhang, Sun, Duvenaud, and Grosse]{zhang2018noisy}
G.~Zhang, S.~Sun, D.~Duvenaud, and R.~Grosse.
\newblock Noisy natural gradient as variational inference.
\newblock In \emph{International conference on machine learning}, pages 5852--5861. PMLR, 2018.

\bibitem[Zhang et~al.(2020)Zhang, Li, Zhang, Chen, and Wilson]{zhang2020csgmcmc}
R.~Zhang, C.~Li, J.~Zhang, C.~Chen, and A.~G. Wilson.
\newblock Cyclical stochastic gradient mcmc for bayesian deep learning.
\newblock \emph{International Conference on Learning Representations}, 2020.

\end{thebibliography}
